\documentclass[11pt,twoside]{article}
\usepackage{amsmath,amsfonts}
\usepackage{algorithmic}
\usepackage{algorithm}
\usepackage{array}
\usepackage[caption=false,font=normalsize,labelfont=sf,textfont=sf]{subfig}
\usepackage{textcomp}
\usepackage{stfloats}
\usepackage{url}
\usepackage{verbatim}
\usepackage{graphicx}
\usepackage{cite}
\usepackage{amsmath,amsfonts,amssymb,cases,mathdots,color,colordvi,url,tikz}
\usepackage{hyperref}
\usepackage{bm}
\usepackage{bbm}
\usepackage{pgfpages}
\usepackage{tabularx}
\usepackage{graphics}
\usepackage{epsfig}
\usepackage[figuresright]{rotating}
\usepackage{rotating}
\usepackage{algorithm}
\usepackage{algorithmic}
\usepackage{physics}
\usepackage{epstopdf}
\usepackage{parskip}
\usepackage{footmisc}
\usepackage{tablefootnote}
\usepackage[justification=centering]{caption}
\usepackage{xcolor}
\usepackage{caption}
\usepackage[normalem]{ulem}

\newtheorem{proposition}{Proposition}[section]
\newtheorem{theorem}[proposition]{Theorem}
\newtheorem{lemma}[proposition]{Lemma}
\newtheorem{corollary}[proposition]{Corollary}
\newtheorem{definition}[proposition]{Definition}

\newtheorem{remark}{Remark}
\newtheorem{assumption}{Assumption}

\newcommand{\be}{\begin{equation}}
	\newcommand{\ee}{\end{equation}}
\newcommand{\ba}{\begin{eqnarray}}
	\newcommand{\ea}{\end{eqnarray}}
\newcommand{\bas}{\begin{eqnarray*}}
	\newcommand{\eas}{\end{eqnarray*}}

\graphicspath{ {./Figures/} }

\def\D{{\mathcal D}}
\def\mbF{{\mathbb F}}

\def\H{{\mathcal{H}}}
\def\T{{\mathbb T}}
\def\mbR{{\mathbb R}}
\def\Re{{\mathbb R}}

\def\S{{\mathcal S}}

\def\F{{\mathcal F}}
\def\I{{\mathcal I}}

\def\N{{\mathcal N}}

\def\T{{\mathcal T}}

\def\bfs{{\bf s}}
\def\bft{{\bf t}}

\def\bfc{{\bf c}}

\def\bfv{{\bf v}}
\def\bfx{{\bf x}}
\def\bfy{{\bf y}}
\def\bfz{{\bf z}}
\def\bfw{{\bf w}}
\def\bfu{{\bf u}}

\def\bfone{{\bf 1}}

\def\bp{ \textbf{Proof.} }
\def\ep{ \hfill $\Box$ }

\def\mbN{ \mathbb{N} }

\def\l01{\ell_{0/1}}
\def\bfal{{\bf {\boldsymbol \alpha}}}

\def\oT{ \overline{T} }

\def\Ga{\Gamma}

\def\argmin{\mathop{{\rm argmin}}}
\def\veps{\varepsilon}

\def\wbz{\widetilde{\bfz}}
\def\wbc{\widetilde{\bfc}}

\def\bfxi{{\boldsymbol \xi}}
\def\bfzt{{\boldsymbol \zeta}}

\title{Local Duality for Sparse Support Vector Machines}

\author{Penghe Zhang\thanks{Department of Data Science and Artificial Intelligence, The Hong Kong Polytechnic University, Hong Kong SAR, China, E-mail: {penghe.zhang@polyu.edu.hk} },
\ \ Naihua Xiu\thanks{School of Mathematics and Statistics, Beijing Jiaotong University, Beijing 100044, China, E-mail: {nhxiu@bjtu.edu.cn} } \ \ and \ \
Houduo Qi\thanks{Department of Data Science and Artificial Intelligence, and Department of Applied Mathematics, The Hong Kong Polytechnic University, Hong Kong SAR, China, E-mail: {houduo.qi@polyu.edu.hk} }
	}

\date{ }

\begin{document}
	
	\maketitle	

	\begin{abstract}
	Due to the rise of cardinality minimization in optimization, sparse support vector machines (SSVMs) have attracted much attention lately and show certain empirical advantages over convex SVMs. A common way to derive an SSVM is to add a cardinality function such as $\ell_0$-norm to the dual problem of a convex SVM. However, this process lacks
	theoretical justification. This paper fills the gap by developing a local duality theory
	for such an SSVM formulation and exploring its relationship with the hinge-loss SVM (hSVM) and the ramp-loss SVM (rSVM). 
	In particular, we prove that the derived SSVM is exactly the dual problem of the 0/1-loss SVM, and the linear representer theorem holds for their local solutions. 
	The local solution of SSVM also provides guidelines on selecting hyperparameters of hSVM and rSVM. {Under specific conditions, we show that a sequence of global solutions of hSVM converges to a local solution of 0/1-loss SVM. Moreover, a local minimizer of 0/1-loss SVM is a local minimizer of rSVM.} This explains why a local solution induced by SSVM outperforms hSVM and rSVM in the prior empirical study. 
	We further conduct numerical tests on real datasets and demonstrate potential advantages of SSVM by working with locally nice solutions proposed in this paper.
	
	\vspace{3mm}
	

\end{abstract}
{}

\section{Introduction}
Significant advances in both hardware technology and software implementation make it possible to 
solve problems previously thought intractable. 
Cardinality minimization, which optimizes functions of counting nonzero elements of variables,
benefits a lot from such advances
as evidenced in a recent survey \cite{tillmann2024cardinality} as well as specific examples in \cite{nikolova2013description,akkaya2020minimizers,akkaya2025minimizers}. The demand for cardinality minimization naturally arises in the support vector machine (SVM, \cite{cortes1995support,vapnik1998statistical,cristianini2000introduction,steinwart2008support,campbell2011learning}), leading to an impressive number of works under the name of Sparse SVM (SSVM). One majority class of SSVM is designed for feature selection, where the redundant features can be removed to reduce computational cost and enhance model interpretability. These SSVMs are often constructed by enforcing sparsity-induced functions such as $\ell_0$-norm \cite{weston2003use,guan2009mixed} and its continuous relaxation  \cite{fung2002minimal, chan2007direct, subrahmanya2009sparse,shao2018sparse, bomze2025feature, zhang2025sparse} on primal SVMs to achieve cardinality minimization of the solution (i.e. normal vector of the decision hyperplane). Another important class of SSVM induces sparsity in the dual variable. Traditionally,
the majority of implementations of convex SVMs are based on computing a dual solution
$\bfal = (\alpha_1,\cdots,\alpha_m)^\top$, and then according to the perfect convex duality theory, the primal solution $\bfw$ can be recovered by the well-known linear representer theorem \cite{dinuzzo2012representer}:
\begin{equation} \label{linear-representation}
	\bfw = \sum_{i=1}^m \alpha_i \bfy_i \bfx_i,
\end{equation}
where $\bfx_i \in \mbR^n$ and $y_i \in \{-1,1\}$ are the $i$th sample and its label respectively for $i = 1,\cdots,m$.  Particularly, the sample vectors $\bfx_i$ which correspond to nonzero $\alpha_i$ are named support vectors. The dual problem is quadratic programming involving an $m \times m$-order dense kernel matrix, which poses significant computation and storage challenges as the scale of data increases. To overcome this drawback, an effective approach is exploiting cardinality minimization of the dual variable (i.e. number of support vectors). Although efficient algorithms have been developed for this class of SSVM (e.g. \cite{zhou2021sparse,wang2021support}), there is lacking theoretical justification, especially for the validity of representer theorem in the nonconvex and sparse setting. This paper aims to consolidate the theoretical foundation for the second class of SSVM. Next, let us review the related works along this line. 
\subsection{Literature review and motivation}

There are many good books on the theoretical and algorithmic aspects of SVMs, see e.g., \cite{vapnik1998statistical, cristianini2000introduction, steinwart2008support, campbell2011learning}.
Of particularly relevant to us is \cite[Sect.~8.4]{steinwart2008support},
where it investigates the sparsity of the dual solution of the hinge-loss SVM (hSVM) by deriving the lower bound of support vectors. A more evident result given in \cite{girosi1998equivalence} is that for noiseless data, the dual hSVM yields the same solution as the sparse approximation problem with $\ell_1$-norm regularization. The analysis of these results largely depends on the (perfect) duality theory in convex optimization. 
However, the dual solution of hSVM tends to lack sufficient sparsity for large-scale or noisy data, which 
would incur higher 
computational cost. As shown in \cite{steinwart2003sparseness-nips}, the number of support vectors grows linearly with the sample size. Moreover, the experimental results in \cite{huang2009arbitrary} indicate that the hSVM generates many irrelevant support vectors when there is a high rate of misclassification. 

To further exploit the sparsity of a dual solution of SVMs, an intuitive approach is enforcing a sparsity-induced $\ell_0$-norm or its approximation on the dual variable. Most prior works along this line adopt more tractable convex or continuous relaxation of $\ell_0$-norm. In \cite{bi2003dimensionality} and \cite{shao2019joint}, linear representation \eqref{linear-representation} is directly substituted into the primal SVMs and then an $\ell_1$-norm $\| \bfal \|_1$ is added in the objective function to derive a sparse dual solution. With a similar framework, a Bayesian learning technique is used in \cite{huang2009arbitrary,huang2010sparse,lopez2011sparse} to approximate $\ell_0$-norm by a sequential procedure. More recent works \cite{zhou2021sparse,landeros2023algorithms} consider directly using $\ell_0$-norm $\| \bfal \|_0$ to obtain a sparse dual solution. Particularly, the SSVM formulation in \cite{zhou2021sparse} is constructed by adding an $\ell_0$-norm constraint to the dual problem of a convex SVM. In this way, the cardinality of the dual solution is controlled at a low level and efficient algorithms can be developed. These works further use linear representation \eqref{linear-representation} to recover the decision hyperplane with high classification performance. However, after adding the sparsity-induced function, original convex SVMs has been changed to a nonconvex SSVM. The validity of \eqref{linear-representation} in nonconvex and sparse setting 
is often assumed and 
requires further theoretical investigation. 

%
%
%
%
%
%

Instead of directly using $\ell_0$-norm, another series of works 
make use of the empirical fact 
that some nonconvex loss functions implicitly yield solutions with fewer support vectors. {A widely studied case is the ramp-loss SVM (rSVM, see \cite{wu2007robust,brooks2011support,cotter2013learning}).} Although the duality theory has not been well-established, 
extensive numerical results indicate that its primal solution can be represented by much fewer support vectors than that of hSVM \cite{collobert2006trading,ong2013learning}. Actually, the ramp loss is a continuous approximation of the 0/1 loss, which is known to be a natural function to characterize the classification problem \cite{wu2007robust,vapnik1998statistical,li2007optimizing}. Traditionally, the 0/1-loss SVM is thought to be intractable, but recently a fast algorithm \cite{wang2021support} is designed for solving 0/1-loss SVM by using the proximal operator of 0/1 loss. 
Its numerical result demonstrates that 0/1-loss SVM not only yields a smaller number of support vectors than that of rSVM but also has competitive classification accuracy compared with hSVM. 
These results 
suggest that 0/1-loss SVM may have certain relevance with SSVMs, rSVM, and hSVM. However, the related study is rare because there lacks duality theory for nonconvex and discontinuous optimization.

\subsection{Main contributions}
The literature review above shows that both $\ell_0$-norm induced SSVM and 0/1-loss SVM lead to solutions with fewer support vectors. The associated algorithms gain significant computational benefits from data reduction. However, two questions are not well addressed.

The first question is referred to as the {\em optimality question} and
it concerns the optimality of the linear representation \eqref{linear-representation} under the sparse setting. We recall that given a dual global solution of a convex SVM, the associated primal global solution can be recovered by \eqref{linear-representation}. However, $\ell_0$-norm leads to nonconvex optimization. Given a solution of such an SSVM, no duality results are available for interpreting the recovered solution from \eqref{linear-representation}. The second question is referred to as the {\em relevance question} and 
it is about the relevance of SSVM, 0/1-loss SVM, rSVM, and hSVM. This question arises from the following facts. 
Firstly, both the $\ell_0$-norm based SSVM and 0/1-loss SVM demonstrate a significant effect on reducing the number of support vectors. We wonder whether there exists a certain relation between them. 
Secondly, since these two problems are nonconvex, the algorithms for them actually find local solutions. However, these solutions demonstrate competitive classification performance when compared with global solutions of hSVM. We hope to provide some theoretical explanation for this phenomenon. 
{Thirdly, the ramp loss is often regarded as a continuous approximation of 0/1 loss, but the relationship between local minimizers of ramp-loss and 0/1-loss SVMs is unclear.}

It turns out that the stationary duality recently proposed by \cite{zhang2025composite} for a class of cardinality optimization plays a pivotal role in answering these two questions. Through the use of this property, we show that the dual problem of 0/1-loss SVM is exactly an $\ell_0$-regularized SSVM formulation. We further characterize the primal-dual solutions and research their relation with hSVM and rSVM. 
In the review above, several SVM models are discussed.
For instance, SSVM largely refers to any SVM model that produces sparse support vectors.
From now on, it mainly refers to the dual problem of $0/1$-loss SVM.
To avoid any confusion, Table~\ref{Table-Abbreviation} lists the abbreviations of the SVM models used in this paper.
\begin{table}[H]
	\caption{Abbreviations of SVMs}
	\label{Table-Abbreviation}
	\centering
		\begin{tabular}{cc}
			\hline
			Abbreviation & Model (to be defined in Sect.~\ref{Section-SVMs})             \\ \hline
			0/1-loss SVM & primal problem of 0/1-loss SVM \eqref{P01} \\
			SSVM         & dual problem of 0/1-loss SVM \eqref{D01}   \\
			hSVM         & hinge-loss SVM \eqref{P1}                  \\
			rSVM         & ramp-loss SVM \eqref{Pr}                   \\ \hline
		\end{tabular}
\end{table}
We summarize the main findings below.
\begin{itemize}
	\item[(i)] Firstly, the local primal and dual solutions of 0/1-loss SVM are characterized in terms of global solutions of hard-margin
	SVM and its dual problem. This allows the convex optimization theory to
	be used. 
	Secondly, given a sparse solution of SSVM which is $\ell_0$-norm driven, we prove that
	the linear representation of the support vectors  is a 
	local optimal solution of the 0/1-loss SVM (i.e., primal local optimal).
	This settles the optimality question and the relevance of 0/1-loss SVM and SSVM discussed above. It can be
	regarded as a generalization of the representation theorem in convex SVMs. 
	Thirdly, such constructed primal and dual pair has a nice property that
	the smooth terms in the objective functions of the two problems are equivalent. This can be regarded as a weak version of the duality of convex SVMs. We refer to such a pair of solutions as {\em locally nice solutions}, see Theorem~\ref{thm-strong-duality-p0}.
	
	\item[(ii)] Given a pair of locally nice solutions, we show that the local solution of 0/1-loss SVM is a limit of global solutions of hinge-loss SVM (hSVM) with suitably selected hyperparameters,
	see Theorem~\ref{thm-p0-sequential-p1}. 
		Moreover, the local minimizer of 0/1-loss SVM is also a local minimizer of rSVM for appropriate hyperparameters, see Theorem \ref{theorem-P01-Pr}. This explains why local solutions of 0/1-loss SVM have competitive classification accuracy relative to hSVM and rSVM.

	\item[(iii)] Empirical study is conducted on real datasets. By using a few evaluation metrics, we show that locally nice solutions yield better performance than other global solutions of hSVM and local solutions of rSVM nearby. 
	This justifies the empirical observation that 0/1-loss SVM and SSVM often perform 
	satisfactorily and lead to better performance than hSVM and rSVM \cite{wang2021support, zhou2021sparse}.
	
\end{itemize}

In our investigation, we made no effort in extending the linear representer 
\eqref{linear-representation} to kernels though it can be done with appropriate 
modifications.
The paper is organized as follows.
In the next section, we review two convex SVMs and the ramp-loss SVM.
They also serve as preparation for introducing the 0/1-loss SVM and its dual problem with the new concept of locally nice solutions.
In Section \ref{Section-Nice-Solutions}, we study the new solution concept, its 
characterizations and present our main results to address the
optimality question. 
We investigate the implications of those characterizations to
the hinge-loss SVM in Section~\ref{Section-Implication}. We research the relationship between 0/1-loss and ramp-loss SVMs in Section \ref{Section-ramp}.
We demonstrate the numerical advantage of the locally nice solutions in Section~\ref{Section-Numerical} and conclude the paper in Section~\ref{Section-Conclusion}.

{\bf Notation:}
We let $\Re^n$ denote the $n$-dimensional Euclidean space with the standard
dot product $\langle \cdot, \cdot \rangle$ and $\mbR^n_+$ is the nonnegative orthant. 
The induced Euclidean norm is $\| \cdot\|$.
The bold-face letters such as $\bfu \in \Re^n$ denote a vector and its $i$th element is denoted by $u_i$.
For a given vector $\bfu \in \Re^n$ and $\varepsilon>0$, the neighborhood centered
at $\bfu$ with radius $\varepsilon$ is denoted as 
$
\N(\bfu, \varepsilon) :=\{
\bfv \; | \; \| \bfu - \bfv\| \le \varepsilon
\},
$ 
where ``$:=$'' means
``define''.
For two vectors $\bfu, \bfv \in \Re^n$, we use $\bfu \perp \bfv$ to say they are 
orthogonal to each other, i.e., $\langle \bfu, \bfv \rangle =0$.
For integer $m>0$, we use $[m]:= \left\{ 1, \ldots, m\right\}$.
For a vector $\bfu \in \Re^m$ and an index set $\I \subset [m]$, $\bfu_\I$ is the subvector of consisting of
components in $\bfu$ indexed by $\I$. The complementary set of $\I$ is denoted by $\overline{\I}$. Given a set $\Theta$, the associated indicator function $\delta_{\Theta}: \mbR^n \to [-\infty, \infty]$ is defined by
\begin{align} \label{ind-fun}
	\delta_{\Theta} (\bfx) := \begin{cases}
		0, &\mbox{if} \ \bfx \in \Theta, \\
		\infty, &\mbox{otherwise}.
	\end{cases}
\end{align} 
Finally, for a vector $\bfal \in \Re^m$, its zero-norm
$\| \bfal\|_0$ counts the number of nonzero elements in $\bfal$.
\section{Four Classes of SVMs} \label{Section-SVMs}

In this part, we review four well-known SVMs:
the hard-margin SVM (and its dual problem), 
the soft-margin SVM (and its dual problem),
the $0/1$-loss SVM (and its dual problem),
and the ramp-loss SVM. The first two are convex problems.
For extensive coverage of SVMs, we refer to 
\cite{cortes1995support, vapnik1998statistical, shawe2004kernel, campbell2011learning} for a few out of many excellent resources.
Suppose we are given data $\{ (\bfx_i, y_i) \}_{i\in [m]}$, where
$\bfx_i \in \Re^n$ are data points and $y_i \in \{-1, 1\}$ are the associated labels.
The purpose is to construct an optimal hyperplane $\H :=\{
\bfx \in \Re^n | \ \langle \bfw, \bfx \rangle + b = 0
\}$
so that it separates the data into two groups according to their labels
with a good generalization property to new data. Here, $\bfw$ is the normal vector of the
hyperplane and $b \in \Re$ is known as the bias.
The normal vector $\bfw$ is often a linear combination of the support
vectors and hence such separation is called the support vector machine. For simplicity, we denote the primal variables $\bfz:=[\bfw;b]$ and $\bfz^*:=[\bfw^*; b^*]$  (Matlab notation for column vectors).

\subsection{Hard-margin SVM and its dual problem}

Most of the standard references for SVMs introduce this case first:  the data $\{(\bfx_i, y_i)\}_{i\in [m]}$ are separable and it can be best modeled by the hard-margin SVM:
\begin{equation} \tag{$P_0$} \label{hm-SVM}
	\begin{aligned}
		\min_{ \bfw, b } & \ \frac{1}{2} \| \bfw \|^2 , \\
		\mbox{s.t.}& \
		y_i (\langle \bfw, \bfx_i \rangle + b) \ge 1 , \quad i \in [m].
	\end{aligned}
\end{equation}
Its dual problem (in the form of minimization) is \cite{vapnik1998statistical, campbell2011learning}:
\begin{equation}\tag{$D_0$} \label{Dual-hm-SVM}
	\begin{aligned}
		\min_{ \bfal \in \Re^m} & \ G(\bfal) := \frac 12 \bfal^\top Q \bfal - \langle \bfone, \bfal \rangle \\
		\mbox{s.t.} & \
		\langle \bfy, \bfal \rangle =0, \ \ \bfal \ge 0 ,
	\end{aligned}
\end{equation}
where $Q_{ij} =y_iy_j \langle \bfx_i, \bfx_j \rangle$ for $i,j \in [m]$, vector $\bfy:= (y_1,\cdots,y_m)^\top \in \mbR^m$, and \textbf{1} is the vector of all ones.
Basic duality requires that $\| \bfw^*\|^2/2 = - G(\bfal^*)$
and $\bfw^* = \sum_{i\in [m]} \alpha^*_i y_i \bfx_i$ (linear representation), where
$\bfw^*$ and $\bfal^*$ are respectively the optimal solutions of 
\eqref{hm-SVM} and \eqref{Dual-hm-SVM}.

\subsection{Soft-margin SVM and its dual problem}

For the more realistic case that the data are not separable, a soft-margin SVM is
constructed. 
A typical one is based on the hinge-loss function (hSVM):
\begin{equation} \tag{$P_1$} \label{P1}
	\begin{aligned}
		&\min_{ \bfz = [\bfw; b] }  F(\bfz):= \notag \\  & \frac{1}{2} \| \bfw \|^2 + \sum_{i = 1}^m c_i \ell_{h} \big( 1 - y_i (\langle \bfw, \bfx_i \rangle + b) \big), 
	\end{aligned}
\end{equation}  
where $c_i \ge 0$ is the weight for the $i$th data point  and
the hinge loss is $\ell_h(t) := \max\{t, 0\}$. 
The hinge loss can be interpreted as $\ell_1$ norm of the slack variables, 
see \cite{campbell2011learning}.
The model \eqref{P1} is often known as weighted SVM
with the weight vector $\bfc = (c_1, \ldots, c_m)$, see \cite{burges1999uniqueness, lapin2014learning}.

The primal problem \eqref{P1} is often solved via its dual problem:
\begin{equation}
	\tag{$D_1$} \label{DualProblem}
	\begin{aligned}
		\min_{\bfal \in \Re^m}& \ G (\bfal) \\ 
		\mbox{s.t.}& \ \ \langle \bfy, \bfal \rangle = 0, \
		0 \le \alpha_i \le c_i, \ i \in[m] .
	\end{aligned}
\end{equation}
If $\bfal^*$ is optimal, then the optimal $\bfw^*$ has a linear representation of the support vectors $\bfx_i$ with $\alpha^*_i >0$:
\be \label{LinearRepresentation}
\bfw^*= \sum_{i \in [m]} \alpha^*_i y_i \bfx_i,
\ee
and the optimal $b^*$ can be calculated using the data points with $0 < \alpha_i^* <c_i$.
We call $(\bfz^*, \bfal^*)$ a primal-dual pair.
Moreover, the convex duality theory ensures $F(\bfz^*) = - G(\bfal^*)$ meaning
that there is no duality gap.

\subsection{SVM with $0/1$ loss and its dual problem}

SVM with $0/1$ loss is traditionally regarded as the ``ideal'' model \cite{vapnik1998statistical}:
\begin{equation}\tag{$P_{0/1}$} \label{P01}
	\min_{ \bfw, b } \frac{1}{2} \| \bfw \|^2 + \lambda \sum_{i = 1}^m \ell_{0/1} \big( 1 - y_i (\langle\bfw,\bfx_i \rangle + b) \big),
\end{equation}
where the hyperparameter $\lambda > 0$ and the zero-one loss function 
$\ell_{0/1}(t) =1$ for $t>0$ and $0$ otherwise.
Due to the counting nature of the $0/1$ loss, the optimization problem is 
regarded as intractable and hence various approximated convex models were proposed.
The hinge-loss SVM \eqref{P1} is one of the earliest such convex modes.
We note that the 0/1-loss SVM treats all data points equally and 
has the same weight $\lambda>0$, while
the hSVM can assign different weights to data points.

We now introduce a dual problem that is based on a newly proposed concept of stationary duality \cite{zhang2025composite}.
For $\bfz =[\bfw; b]$, $\bfs, \bft \in \Re^m$ and $\mu >0$, let
\begin{align}
	& f(\bfz) := \frac 12 \| \bfw\|^2, \quad
	\varphi(\bft) := \lambda \sum_{i=1}^m \ell_{0/1} (t_i) \notag\\
	& g(\bfs) := \mu \| \bfs\|_0 + \delta_{\Re^m_+}(\bfs), \ 
	A := - \begin{bmatrix}
		y_1 \bfx_1^\top, \ & y_1 \\
		\vdots \ & \vdots \\
		y_m \bfx_m^\top , \ & y_m
	\end{bmatrix}, \label{data-A}
\end{align}
where $\mbR^m_+$ is the nonnegative orthant. Both $\varphi(\cdot)$ and $g(\cdot)$ are nonsmooth and can be separated into the sum of $m$ univariate functions. For simplicity, let us consider the case $m = 1$. Then their limiting subdifferentials \cite{RockWets98} can be
characterized as
\[
\partial \varphi (t) = \left\{
\begin{array}{ll}
	0, &            \mbox{if} \ t \not=0 \\
	\mbox{[}0, \infty), & \mbox{if} \ t =0 
\end{array} 
\right .
\]
\[
\partial g (s) = \left\{
\begin{array}{ll}
	\{0\}, &            \mbox{if} \ s >0 \\
	\Re, & \mbox{if} \ s =0 
\end{array} 
\right . \ \mbox{for} \ s\ge 0.
\]
and another important property about $\varphi(\cdot)$ and $g(\cdot)$ is that
\[
s \in \partial \varphi(t) \quad 
\Longleftrightarrow \quad
t \in \partial g(s).
\]
This property is called the {\em stationarity symmetry} property in \cite{zhang2025composite}
and can be extended componentwise to the $m$-dimensional case of $\varphi(\bft)$ and
$g(\bfs)$ for $\bfs, \bft \in \Re^m$.
Problem \eqref{P01} can be reformulated as
\[
\min_{\bfz} \ f(\bfz) + \varphi(A\bfz + \bfone).
\]
Its stationary dual problem is given by (in the minimization form):
\[
\min_{\bfal \in \Re^m}\ f^*(-A^\top \bfal) - \langle \bfone, \bfal \rangle + g(\bfal),
\]
	where $f^*: \mbR^{n+1} \to (- \infty, \infty]$ is the conjugate function (see \cite{rockafellar1970convex}) of $f$, taking the following representation:
	\begin{align*}
		f^*(\bfzt) :=& \max_{\bfz = [\bfw;b]} \langle \bfzt, \bfz \rangle - f(\bfz) \\
		=& \max_{\bfz = [\bfw;b]} \sum_{i = 1}^n (w_i \zeta_i - \frac{1}{2} w_i^2) + \zeta_{n+1} b \\
		=& \frac{1}{2} \sum_{i = 1}^n \zeta_i^2 + \delta_{\{ 0 \}}(\zeta_{n+1}).
	\end{align*}
Making use of the structure in the matrix $A$ in \eqref{data-A}, we have
\[
A^\top \bfal = -\sum_{i=1}^m \alpha_i \left[
\begin{array}{c}
	y_i \bfx_i \\
	y_i
\end{array} 
\right] = - \left[ 
\begin{array}{c}
	\sum_{i=1}^m \alpha_i	y_i \bfx_i \\
	\sum_{i=1}^m \alpha_i	y_i
\end{array} 
\right].
\]
Substituting it to $f^*(-A^\top \bfal )$ yields the dual problem
below:
\begin{equation}\tag{$D_{0/1}$} \label{D01}
	\begin{aligned}
		\min_{ \bfal \in \Re^m } & \ \ G_0(\bfal) := G(\bfal) + \mu \| \bfal\|_0 \\
		\mbox{s.t.} & \ \ \langle \bfy, \bfal \rangle = 0,
		\quad \bfal \geq 0,
	\end{aligned} 
\end{equation}
where $\mu >0$ and the $\ell_0$-norm of $\bfal$ is due to $g(\bfs)$.

The stationary dual problem \eqref{D01} is not studied before and
can be regarded as an $\ell_0$-norm regularized SSVM.
Our main task below is to show that its local solution yields an
optimal solution of \eqref{P01} that can be represented as a
sparse representation of its support vectors. 

	\begin{remark}
		We note that a widely studied hSVM model \cite{cortes1995support} has just one positive parameter $c$, corresponding to $c_i=c$ for all $i$ in the vector weights $\bfc$ in \eqref{P1}.
		The weighted SVM \eqref{P1} has also been extensively studied for applications such as the cost-sensitive classifications, see e.g., 
		\cite{burges1999uniqueness,lapin2014learning,karasuyama2012multi,iranmehr2019cost}.
		It is the weighted model that enjoys a solution property closely related to the
		0/1-loss SVM, see {Theorem~\ref{thm-p0-sequential-p1}} below.
		We emphasize that this solution property is unlikely to hold for the hSVM with a single parameter $c$.
		
		
	\end{remark}
	\subsection{SVM with the ramp loss}
	Given $\gamma > 0$, the ramp loss $\ell_\gamma: \mbR \to \mbR$ is defined as
	\begin{align*}
		\ell_\gamma (t):= \left\{ \begin{aligned}
			& 1, && \mbox{if}~ t > \gamma, \\
			& t, && \mbox{if}~ 0 \leq t \leq \gamma,\\
			& 0, && \mbox{if}~ t < 0.
		\end{aligned} \right.
	\end{align*}
	As a continuous approximation of 0/1 loss, it helps improve the robustness to outliers and reduce the number of support vectors \cite{wu2007robust,brooks2011support,cotter2013learning,collobert2006trading}. The ramp loss SVM (rSVM) is as follows:
	\begin{equation}\tag{$P_{r}$} \label{Pr}
		\min_{ \bfw, b } \frac{1}{2} \| \bfw \|^2 + \rho \sum_{i = 1}^m \ell_{\gamma} \big( 1 - y_i (\langle\bfw,\bfx_i \rangle + b) \big),
	\end{equation}
	where $\rho > 0$ is the penalty parameter. To the best of our knowledge, there is no well-established dual formulation of the nonconvex optimization \eqref{Pr}. Therefore, we focus on the primal optimization \eqref{Pr} in this work. 

Before we proceed with our technical proofs, let us explain the notation used for the above SVMs. 
We label the hard-margin SVM as \eqref{hm-SVM} because no
misclassification occurs for the separable case. We label the soft-margin SVM as \eqref{P1} because it can be interpreted as $\ell_1$-norm
loss. Obviously, \eqref{Pr} and \eqref{P01} are because of ramp loss and the $0/1$ loss used.
Their corresponding dual problems are hence denoted as \eqref{Dual-hm-SVM},
\eqref{DualProblem}, and \eqref{D01} respectively.

\section{Locally Nice Solutions} \label{Section-Nice-Solutions}

This section aims to address the optimality problem discussed in the Introduction for SSVM.
The answer is embedded in the 
	\emph{``\textbf{locally nice solutions}"},
a
new solution concept to be introduced first.
We then characterize the primal and dual local solutions of
\eqref{P01} and \eqref{D01} by the global solutions of the corresponding hard-margin SVM.
Finally, we match the primal and dual solutions as a locally nice pair. 

	The optimality conditions of $\ell_0$-norm regularized least square loss, Huber loss, and least absolute deviation have been comprehensively studied in \cite{nikolova2013description,akkaya2020minimizers,akkaya2025minimizers} respectively. These papers and the references therein provide readers with a deeper understanding on the sparsity-regularized optimization problems. However, we need to stress that problems \eqref{P01} and \eqref{D01} have additional composite structure or constraints compared with these problems. They require more specialized investigation in this section.

\subsection{Locally nice solutions}

Our conceptual methodology for addressing the optimality problem
is as follows.
We start with a local solution $\bfal^*$ of the dual problem \eqref{D01}.
We then construct the normal vector $\bfw^*$ as in \eqref{LinearRepresentation} to be a linear combination of the support vectors and we prove the optimality of $\bfw^*$ as a local solution
of the primal problem \eqref{P01}.
This is usually true for convex SVMs. 
However, this cannot be directly extended to each pair of local optimal solutions of 0/1-loss SVM 
\eqref{P01} and its dual \eqref{D01}. 
Let us consider the
XOR (exclusive-or) data with $\bfx_i  \in \{(0;0), (1;0), (1;1), (0;1)\}$ and the
corresponding label $y_i \in \{1, -1, 1, -1\}$.
We can verify that $\bfw^* = (0,-2)$ and $b^*=1$ is a local solution
of \eqref{P01} (by Lemma \ref{Lemma-Characeterization-Local-Solutions}). 
On the other hand,
$\bfal^*=(0,2,2,0)$ is a local solution of the dual problem
\eqref{D01} (by Lemma \ref{Lemma-Dual-Solutions}). Apparently, the linear representation 
\eqref{LinearRepresentation} does not hold for $\bfw^*$ and $\bfal^*$. This motivates us to propose the following important solution
concept. 

\begin{definition}[Locally nice pair] \label{Def-Nice-Solutions}
	We say a primal-dual pair $(\bfz^*, \bfal^*)$ is locally nice
	with respect to the primal and dual problem \eqref{P01} and \eqref{D01} 
	if the following conditions hold:
	\be \label{Nice-Conditions}
	\left\{
	\begin{aligned}
		& \bfz^* \ \mbox{is a local solution of} \ \eqref{P01}, \\
		&	\bfal^* \ \mbox{is a local solution of} \ \eqref{D01}, \\
		& \bfw^* = \sum_{i=1}^m \alpha^*_i y_i \bfx_i \  \mbox{and} \
		\| \bfw^*\|^2/2 = -G(\bfal^*) .
	\end{aligned}
	\right .
	\ee  
\end{definition}
We note that the linear representation is part of the necessary condition for a
pair of solutions to be nice.
Furthermore, we require the objectives to be equal $\| \bfw^*\|^2/2 = -G(\bfal^*)$.
This is different from the duality for hSVM where $F(\bfz^*) = - G(\bfal^*)$.
We note that $\| \bfw\|^2/2$ is the smooth part of $F(\bfz)$
and $G(\bfal)$ is the smooth part of $G_0(\bfal)$.
That is,  
we only require the smooth parts of the objectives to be equal at local solutions.
It is easy to see locally nice pair automatically gives rise to the optimality of 0/1-loss SVM and SSVM.
Our main task is to show that such a pair always exists.
\subsection{Characterization of primal solutions}

We consider that there are at least two data points with opposite labels in the data $\{(\bfx_i, y_i)\}_{i \in [m]}$. Let $\I$ be a fixed index subset of $[m]$ consisting of the data points $\{(\bfx_i, y_i)\}_{i \in \I}$ that are separable. Then the hard-margin SVM is well-defined on such data:
\be \tag{$P_{\I}$}\label{hm-SVM-I}
\begin{aligned}
	\min_{ \bfw, b }& \ \frac{1}{2} \| \bfw \|^2 ,\\
	\mbox{s.t.} & \
	y_i (\langle \bfw, \bfx_i \rangle + b) \ge 1 , \quad i \in \I.
\end{aligned}
\ee 
Next, we will show that a local solution of 0/1-loss SVM \eqref{P01} can be characterized as a global solution of \eqref{hm-SVM-I} with certain choices of $\I$. Before that, let us define some index sets associated with given $\bfz := [\bfw; b]$ and $\bfz^*:= [\bfw^*; b^*]$:
\be \label{Index-I0}
\left\{
\begin{aligned}
	&u_i(\bfz) := 1 - y_i( \langle \bfw,\bfx_i \rangle + b), \ \mbox{for} \ i \in [m],
	\\ 
	&\I_0 (\bfz)  := \left\{ i \in [m] \ | \ u_i(\bfz) \le 0\right\} ,\\
	&	\I_+(\bfz) := \left\{ i \in [m] \ | \ u_i(\bfz) > 0\right\}, \\
	& u^*_i = u_i(\bfz^*), \  
	\I_0^* := \I_0(\bfz^*), \  \I_+^* := \I_+(\bfz^*). 
\end{aligned}
\right .
\ee
Obviously, we have $\I_0(\bfz) \cup \I_+(\bfz) = [m]$. Now we consider the hard-margin SVM \eqref{hm-SVM-I} with $\I = \I^*_0$. We have the following result.

\begin{lemma}\label{Lemma-Characeterization-Local-Solutions}
	The point $\bfz^* = [\bfw^*; b^*]$ is a local solution of 0/1-loss SVM \eqref{P01} 
	if and only if $\bfz^*$ is a global solution of the hard-margin SVM \eqref{hm-SVM-I}
	with $\I = \I_0^*$.	
\end{lemma}

\bp
Suppose $\bfz^*$ is a local solution of \eqref{P01}, there exists $\veps > 0$ such that for any $\bfz \in \N(\bfz^*,\veps)$, the following inequality holds 
\begin{align*}
	&\frac{1}{2} \| \bfw \|^2 + \underbrace{\lambda \sum_{i = 1}^m \ell_{0/1} \big( 1 - y_i (\langle \bfw, \bfx_i \rangle + b) \big)}_{\lambda | \I_+(\bfz) | } 	\geq  \frac{1}{2} \| \bfw^* \|^2 + 
	\underbrace{\lambda \sum_{i = 1}^m \ell_{0/1} \big( 1 - y_i (\langle \bfw^*, \bfx_i \rangle + b^*) \big)}_{\lambda | \I_+^* |}.
\end{align*}
Let us denote the feasible region of \eqref{hm-SVM-I} with $\I=\I^*_0$ as $\F_*$ and
the thus obtained problem as $P_*$.
Then for any $\bfz \in \N(\bfz^*,\veps) \cap \F_*$, we have 
$
| \I_+(\bfz) | \le | \I_+^* |,
$
which leads to $\| \bfw \|^2/2 \geq \| \bfw^* \|^2/2$ for all $\bfz \in \N(\bfz^*,\veps) \cap \F_*$. 
This means that $\bfz^*$ is a local solution of $P_*$.
Since $P_*$ is convex (hard-margin SVM), a local solution is also global. 
This proves that $\bfz^*$ is a global solution of $P_*$.

Conversely,  suppose $\bfz^*$ is a global solution of $P_*$, we have
\begin{align} \label{global-min-S*}
	\| \bfw \|^2/2 \geq \| \bfw^* \|^2/2, \ \ \forall \ \bfz \in \F_*.
\end{align}
Let us take a sufficiently small radius $\veps_1 > 0 $ such that 
\begin{align} \label{small-veps}
	\| \bfw \|^2/2 \geq \| \bfw^* \|^2/2 - \lambda/2 \quad \mbox{and} \quad
	\I_+(\bfz) \supseteq \I_+^*
\end{align}
for any $\bfz \in \N(\bfz^*, \veps_1)$, where the above two inequalities follow from the continuity of the quadratic and linear functions respectively. 
	Considering the formula \eqref{global-min-S*} only holds for $\bfz \in \F_*$, we divide the neighborhood of $\bfz^*$ into $\N(\bfz^*,\veps_1) \cap \F_*$ and $\N(\bfz^*,\veps_1) \cap \overline{\F}_*$. Then let us show that $\bfz^*$ is the minimizer in these two regions.

\underline{If $\bfz \in \N(\bfz^*,\veps_1) \cap \F_*$}, then \eqref{global-min-S*} and the second relationship in \eqref{small-veps} directly imply 
\begin{align*}
	& \frac{1}{2} \| \bfw \|^2 + \lambda \sum_{i = 1}^m \ell_{0/1} \big( 1 - y_i (\langle \bfw, \bfx_i \rangle + b) \big) \geq  \frac{1}{2} \| \bfw^* \|^2 + \lambda \sum_{i = 1}^m \ell_{0/1} \big( 1 - y_i (\langle \bfw^*, \bfx_i \rangle + b^*) \big).
\end{align*}

\underline{If $\bfz \in \N(\bfz^*,\veps_1) \cap \overline{\F}_*$}, then $\bfz \notin \F_*$. This means $\I_+(\bfz)$ contains at least one more element than $\I_+^*$,
i.e., $|\I_+(\bfz)| \ge |\I_+^*| + 1$. Combining this with the first inequality in \eqref{small-veps}, we have
\begin{align*}
	& \frac{1}{2} \| \bfw \|^2 + \lambda \sum_{i = 1}^m \ell_{0/1} \big( 1 - y_i (\langle \bfw, \bfx_i \rangle + b) \big) \\
	\ge & \frac{1}{2} \| \bfw^* \|^2 - \frac{\lambda}2 + \lambda |\I_+(\bfz)| \ge \frac{1}{2} \| \bfw^* \|^2 - \frac{\lambda}2 + \lambda |\I_+^*| + \lambda \\
	> &\frac{1}{2} \| \bfw^* \|^2 + 
	\lambda \sum_{i = 1}^m \ell_{0/1} \big( 1 - y_i (\langle \bfw^*, \bfx_i \rangle + b^*) \big).
\end{align*}
We conclude that \eqref{P01} has the smallest objective function value at $\bfz^*$ in the neighborhood  $\N(\bfz^*,\veps_1)$ of $\bfz^*$. 
Hence, $\bfz^*$ is its local solution.
\ep

\begin{remark} \label{Remark-Primal}
	If we choose $\I$ consisting maximum number of separable data points.
	Then the hard-margin SVM \eqref{hm-SVM-I} has an optimal solution. 
	This solution must be a local solution of \eqref{P01} according to 
	Lemma~\ref{Lemma-Characeterization-Local-Solutions}. 
	%
\end{remark} 

\subsection{Characterization of dual solutions}

The above result suggests that a local solution of the 0/1-loss SVM dual problem
\eqref{D01} may be a global solution of the dual problem of the hard-margin SVM
\eqref{hm-SVM-I} with proper choice $\I$. Next we aim to verify this result. 
Let us recall the dual problem of \eqref{hm-SVM-I}:
\be \tag{$D_\I$} \label{Dual-DI}
\begin{aligned}
	\min_{\bfal \in \Re^m} & \ G(\bfal) \\ \mbox{s.t.} & \
	\bfy^\top \bfal=0, \quad \bfal_{\overline{\I}} = 0, \quad
	\bfal_{\I} \ge 0 .
\end{aligned}
\ee 
We define the following notation.
\[
\left\{
\begin{aligned}
	\F_D & := \{ \bfal \in \mbR^m\; | \; \bfal \geq 0,\ \ \langle \bfy, \bfal \rangle = 0 \}, \\
	T^* & := \left\{ i \in [m] \; | \; \alpha^*_i \not= 0 \right\} \ \ \mbox{for} \
	\bfal^* \in \F_D , \\
	\F^*_D &:= \{ \bfal \in \mbR^m \; | \; \bfal_{T^*} \geq 0,  \bfal_{\oT^*} = 0,  \langle \bfy, \bfal \rangle = 0 \}.
\end{aligned} 
\right .
\]
In other words, $\F_D$ is the feasible region of  the 0/1-loss SVM dual problem
\eqref{D01} and $\F^*_D$ is the feasible region of the dual problem
\eqref{Dual-DI} with $\I = T^*$.
We have the following result.

\begin{lemma} \label{Lemma-Dual-Solutions}
	It holds that $\bfal^*$ is a local solution of the 0/1-loss SVM dual problem 
	\eqref{D01} if and only if it is a global solution of \eqref{Dual-DI} with 
	$\I = T^*$.
\end{lemma} 

\bp
Suppose $\bfal^*$ is a local solution of \eqref{D01}, then there exists $\veps_2 > 0$ such that 
\begin{align*}
	G(\bfal) + \mu \| \bfal \|_0 \geq 
	G(\bfal^*) + \mu \| \bfal^* \|_0
\end{align*}
for any $\bfal \in \N(\bfal^*, \veps_2) \cap \F_D$. Considering $\mu \| \bfal \|_0 \leq \mu \| \bfal^* \|_0 $ holds when $\bfal_{\oT^*} = 0$, we have
\begin{align*}
	G(\bfal) \geq G(\bfal^*), \ \ \forall \bfal \in \N(\bfal^*, \veps_2) \cap \F^*_D.
\end{align*}
This proves that $\bfal^*$ is a local solution of \eqref{Dual-DI} with
$\I=T^*$. Therefore, it is also a global solution since the problem is convex.

Conversely, suppose $\bfal^*$ is a global solution of \eqref{Dual-DI} with
$\I=T^*$ , we have
\begin{align} \label{loc-min-cvx*}
	G(\bfal) \geq G(\bfal^*), \ \ \forall \bfal \in  \F^*_D.
\end{align}
We assume $\veps_3>0$ is sufficiently small such that the following formulas hold for any $\bfal \in \N(\bfal^*, \veps_3)$
\begin{align} \label{neighborhood-prop}
	\begin{aligned}
		&T:= \{ i \in [m] \ | \ \alpha_i \neq 0 \} \supseteq T^*, \\
		&G(\bfal) \geq G(\bfal^*) - \mu/2,
	\end{aligned}
\end{align}
where the second line in the above formula is from the continuity of $G$.
Denoting $\Xi^*:= \{ \bfal \in \mbR^m \ | \ \bfal_{\oT^*} = 0 \}$, and hence $\F^*_D = \F_D \cap \Xi^*$. 
	Since formula \eqref{neighborhood-prop} only holds for $\bfal \in  \F^*_D$, we divide the neighborhood of $\alpha^*$ into $\N(\bfal^*, \varepsilon_3) \cap \F_D \cap \Xi^*$ and $\N(\bfal^*, \varepsilon_3) \cap \F_D \cap \overline{\Xi}^*$. Next let us show that $\alpha^*$ is the minimizer in these two regions.

\underline{When $\bfal \in \N(\bfal^*, \varepsilon_3) \cap \F_D \cap \Xi^*$}, the first line of \eqref{neighborhood-prop} and $\bfal \in \Xi^*$ imply $\mu \| \bfal \|_0 = \mu \| \bfal^* \|_0$. Combining this with \eqref{loc-min-cvx*}, we can derive $G(\bfal) + \mu \| \bfal \|_0 \geq G(\bfal^*) + \mu \| \bfal^* \|_0$.

\underline{When $\bfal \in \N(\bfal^*, \varepsilon_3) \cap \F_D \cap \overline{\Xi}^*$}, the first line of \eqref{neighborhood-prop} implies $\mu \| \bfal \|_0 \ge \mu \| \bfal^* \|_0 + \mu$. This, together with the second line of \eqref{neighborhood-prop}, yields $G(\bfal) + \mu \| \bfal \|_0 \geq G(\bfal^*) + \mu \| \bfal^* \|_0 + \mu/2$.
We proved that $\bfal^*$ is a local solution of  \eqref{D01}.
\ep

\subsection{Matching the primal and dual solutions as locally nice pair}


Suppose $\bfal^*$ is a local solution of \eqref{D01}.
Lemma~\ref{Lemma-Dual-Solutions} shows that it is also a global solution
of the hard-margin SVM dual problem \eqref{Dual-DI} with $\I = T^*$. 
Since \eqref{Dual-DI} is convex, its solution can be equivalently characterized
by the Karush-Kuhn-Tucker (KKT) conditions with $(\bfu^*, b^*)$ as the Lagrange
multipliers: 
\begin{align} \label{KKT-T*}
	\left\{ \begin{aligned}
		& \nabla G (\bfal^*) + \bfu^* + b^* \bfy = 0, \\
		& 0 \leq \bfal^*_{T^*} \perp \bfu^*_{T^*}  \leq 0, \\
		& \bfal^*_{\oT^*} = 0, \ \langle \bfy, \bfal^* \rangle = 0.
	\end{aligned} \right.
\end{align}
We have the following key result.

\begin{theorem} \label{thm-strong-duality-p0} 
	Let $\bfal^*$ be a local solution of the 0/1-loss SVM dual problem 
	\eqref{D01}. We further let $(\bfu^*, b^*)$ be the Lagrange  multipliers satisfying
	\eqref{KKT-T*}. Let 
	$
	\bfw^* := \sum_{i=1}^m \alpha_i^* y_i \bfx_i .
	$
	Then the following hold.
	
	(i) Point $\bfz^* := [\bfw^*; b^*]$ is a local solution of \eqref{P01}.
	
	(ii) It holds that $\| \bfw^* \|^2/2 = -G(\bfal^*)$.
	
	In other words, the thus defined $(\bfz^*, \bfal^*)$ is locally nice. 
\end{theorem}
\bp
(i) We expand the first equation in \eqref{KKT-T*} to get
\[
Q \bfal^* - \bfone + \bfu^* + b^* \bfy = 0.
\]
This gives $\bfu^*= \bfone - (Q \bfal^* + b^* \bfy )$. 
We calculate its component
\begin{align*}
	u^*_i &= 1 - \Big(
	y_i \langle \bfx_i, \sum_{j \in [m]} y_j \alpha^*_j \bfx_j \rangle + b^* y_i
	\Big) = 1 - y_i \Big(   
	\langle \bfx_i, \bfw^* \rangle + b^*
	\Big) , \quad i \in [m] . 
\end{align*} 
We note that $u^*_i \le 0$ for $i \in T^*$. Consequently, we reformulate \eqref{KKT-T*} as
\begin{align} \label{KKT-I*}
	\left\{ \begin{aligned}
		& \bfw^*= \sum_{i=1}^m \alpha_i^* y_i  \bfx_i, \\
		& 1 - y_i ( \langle \bfw^*, \bfx_i \rangle + b^*) \le 0 , \ \alpha_i^* \ge 0, \ i \in T^*, \\
		& \alpha^*_i (1 - y_i ( \langle \bfw^*, \bfx_i \rangle + b^*)) = 0, \ i \in T^*, \\
		& \alpha^*_i = 0, \ \ i \not\in T^*, \quad \langle \bfy, \bfal^* \rangle = 0.
	\end{aligned} \right.
\end{align}
Those are exactly the KKT conditions of the convex problem \eqref{hm-SVM-I} with $\I = T^*$ and $\bfal^*_{T^*}$ being its Lagrange multiplier.
Here comes the crucial observation with $\I^*_0 = \left\{ i \in [m] \; | \; u^*_i \le 0 \right\}$ and $T^* \subseteq \I^*_0$: 
\begin{equation} \label{alpha-u-relation}
	\left\{
	\begin{array}{ll}
		\alpha^*_i	> 0 \ \mbox{and} \ u^*_i \leq 0, & \mbox{for} \ i \in \I_0^* \cap T^*, \\
		\alpha^*_i =0 \ \mbox{and} \ u^*_i \leq 0,   & \mbox{for} \ i \in \I^*_0 \setminus T^*, \\
		\alpha^*_i =0 \ \mbox{and} \ u^*_i > 0 ,   & \mbox{for} \ i \in \I^*_+ . 
	\end{array} 
	\right .
\end{equation}
The conditions in \eqref{KKT-I*} can be restated in terms of $\I^*_0$ as follows:
\begin{align*} 
	\left\{ \begin{aligned}
		& \bfw^*= \sum_{i=1}^m \alpha_i^* y_i  \bfx_i, \\
		& 1 - y_i ( \langle \bfw^*, \bfx_i \rangle + b^*) \le 0 , \ \alpha_i^* \ge 0, \ i \in \I_0^*, \\
		& \alpha^*_i (1 - y_i ( \langle \bfw^*, \bfx_i \rangle + b^*)) = 0, \ i \in \I_0^*, \\
		& \alpha^*_i = 0, \ \ i \not\in \I_0^*, \quad \langle \bfy, \bfal^* \rangle = 0.
	\end{aligned} \right.
\end{align*}
Those conditions are exactly the KKT conditions of convex SVM \eqref{hm-SVM-I} with $\I = \I^*_0$. Hence, $\bfz^*$ is its global solution. 
Lemma~\ref{Lemma-Characeterization-Local-Solutions} says that
$\bfz^* = [\bfw^*; b^*]$ must be a local solution of
0/1-loss SVM primal problem \eqref{P01}.

(ii) Using the KKT conditions \eqref{KKT-T*}, we obtain
\begin{align*}
	& \frac{1}{2} \| \bfw^* \|^2 + G(\bfal^*) \\
	=& \frac{1}{2} {\bfal^*}^\top Q \bfal^* + \frac{1}{2} {\bfal^*}^\top Q \bfal^* - \langle \textbf{1}, \bfal^* \rangle \\
	=& \langle \bfal^*, Q\bfal^* - \textbf{1} \rangle = \langle \bfal^*, -b^*\bfy - \bfu^* \rangle \\
	= & - b^* \langle \bfy, \bfal^* \rangle - \langle \bfu^*, \bfal^* \rangle = 0,
\end{align*}
where the first equality is from $\bfw^* = \sum_{i=1}^{m} \alpha^*_i y_i\bfx_i$, the third equality follows from the first equation in \eqref{KKT-T*}, and the last equality is derived by $\langle \bfy, \bfal^* \rangle = 0$ and $ \bfu^* \perp {\bfal^*}$.
\ep

Theorem \ref{thm-strong-duality-p0} basically indicates that a primal solution of \eqref{P01} can be recovered by a dual solution of \eqref{D01}. They turn out to be a pair of locally nice solutions satisfying the linear representation theorem. The subsequent result reveals that the locally nice solutions can also be obtained through the convex hinge-loss SVM \eqref{P1} with a proper setting of weights $c_i$ for $i \in [m]$.

\begin{corollary} \label{thm-sol-01-H-SVM}
	Let $(\bfz^*, \bfal^*)$ be locally nice solutions, where $\bfz^* = [\bfw^*; b^*]$ was defined as that in Theorem~\ref{thm-strong-duality-p0}. $\I_0^*$ is defined as that
	in \eqref{Index-I0} and we take
	\begin{align} \label{c*1}
		c^*_i \left\{ \begin{aligned}
			&\geq \alpha^*_i, && \mbox{if} \ \ i \in \I_0^*, \\
			& = \alpha^*_i, && \mbox{if} \ \ i \not \in \I_0^*.
		\end{aligned} \right.
	\end{align}
	Then $\bfz^*$ is an optimal solution of \eqref{P1} with $\bfc = \bfc^*$.
\end{corollary}

\bp
Note that $\alpha^*_i=0$ for $i \not\in \I^*_0$ due to \eqref{alpha-u-relation} in Theorem \ref{thm-strong-duality-p0}. The choice of $\bfc = \bfc^*$ means
that Problem \eqref{P1} can be represented as
\[
\min_{ \bfz = [\bfw; b] }\ F(\bfz) = \frac{1}{2} \| \bfw \|^2 + \sum_{i \in \I^*_0} c_i \ell_{h} \big( 1 - y_i (\langle \bfw, \bfx_i \rangle + b) \big) .
\]
The dual optimization of the above problem can be written as
\begin{equation} \label{dual-I0}
	\begin{aligned}
		\min_{\bfal \in \Re^m}& \ G (\bfal) \\ 
		\mbox{s.t.}& \ \langle \bfy, \bfal \rangle = 0, && \\
		&\ 0 \le \alpha_i \le c_i, && \ i \in \I^*_0 \\
		&\ \alpha_i = 0, && \ i \notin \I^*_0
	\end{aligned}
\end{equation}
Particularly, $\bfal^*$ is a feasible point of \eqref{dual-I0}. By weak duality theorem (e.g. \cite[Theorem 12.11]{nocedal2006numerical}), the following inequality holds for any $\bfz \in \mbR^m$:
\begin{equation*}
	F(\bfz) \geq -G(\bfal^*)
\end{equation*}
By the choice $\bfc = \bfc^*$ and definition of locally nice solutions, we can obtain $F(\bfz^*) = \| \bfw^* \|^2/2 = -G(\bfal^*)$. This means that $\bfz^*$ is an optimal solution of \eqref{P1} with $\bfc = \bfc^*$. 
\ep

\begin{remark}
	From Corollary \ref{thm-sol-01-H-SVM}, we know that local solutions of \eqref{P01} actually belong to a subset of the global solutions of \eqref{P1}. Furthermore, to recover these solutions, the local solutions of \eqref{D01} provide guidelines on the selection of weighted parameters of \eqref{P1}. However, this setting requires that $\alpha^*_i =0$ for $i \not\in \I_0^*$. That is, the weight vector contains zero weights.
	This is against the common practice that the weights are often positive, though they
	can be very small. 
	It would be very useful if we can prove that the solutions of the convex hinge-loss SVM
	converges to a locally nice solution when the positive weights converge to $\bfc^*$. 
	This expectation seems reasonable, but it is closely related to the optimization stability theory. We give a positive answer with a formal
	mathematical proof in the next section.
\end{remark}

	\begin{remark} \label{rmk-loc-nice-sol-interp}
		At the end of this section, we would like to give some geometric and statistical interpretation about the locally nice solutions of \eqref{P01}.
		
		(i) Geometric interpretation: Let us consider the one-dimensional samples $\bfx_i \in \{ 0, 1, 2 \}$ with $y_i \in \{ 1,-1,-1 \}$. We take $\lambda = 10$ and $\mu = 1/5$ for \eqref{P01} and \eqref{D01}. It can be verified that there are two pairs of nonzero locally nice solutions: ${\bfz^{(1)}} = (-2;1),{\bfal^{(1)}} = (2;2;0)$ and ${\bfz^{(2)}} = (-1;1),{\bfal^{(2)}} = (1/2;0;1/2)$. 
		Since there exists the constraint $\alpha_1 = \alpha_2 + \alpha_3$ in \eqref{D01}, we may eliminate $\alpha_1$ so that we can visualize the landscape of
		the objectives in terms of $\alpha_2$ and $\alpha_3$. We denote the objective functions of \eqref{P01} and \eqref{D01} as $F_{0/1}$ and $G_0$ respectively. The left column of Fig.~\ref{fig_minimizer} demonstrates the separated pieces of the objective functions of \eqref{P01} and \eqref{D01} with jumps and the right column shows the corresponding domain of piecewise linear regions (i.e., the union of polyhedral regions). On each of the pieces, both of the primal and dual objectives are convex quadratic functions.
		\begin{figure}[htb]
			\subfloat{
				\begin{minipage}[t]{0.5\linewidth}
					\centering
					\includegraphics[width=2.4in]{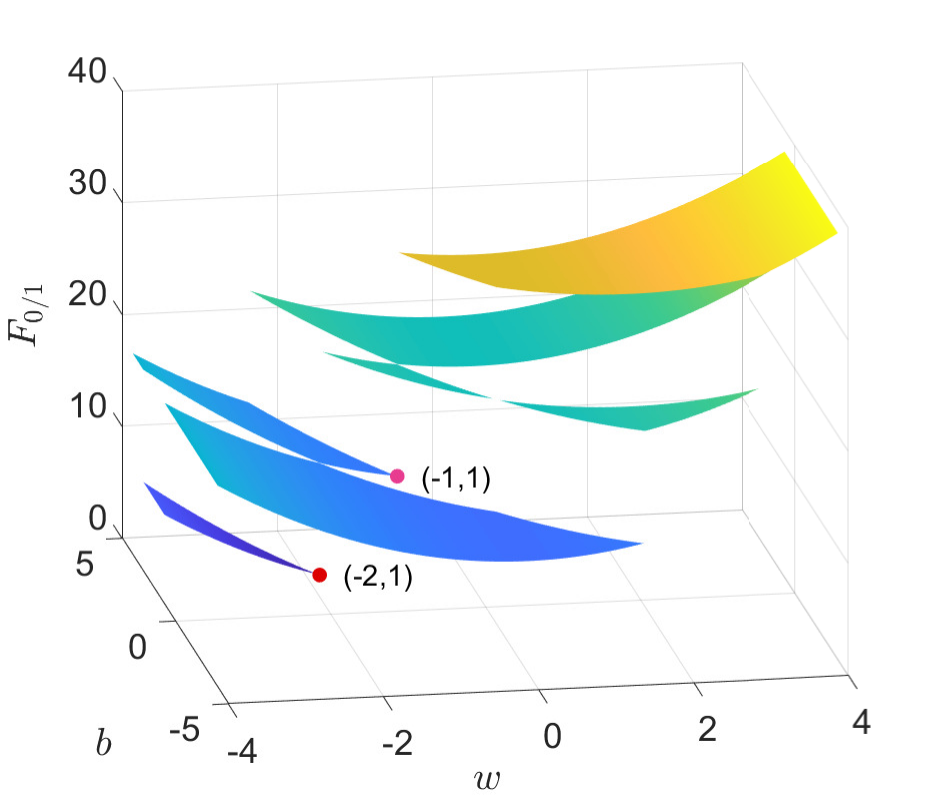}
				\end{minipage}%
				\begin{minipage}[t]{0.5\linewidth}
					\centering
					\includegraphics[width=2.4in]{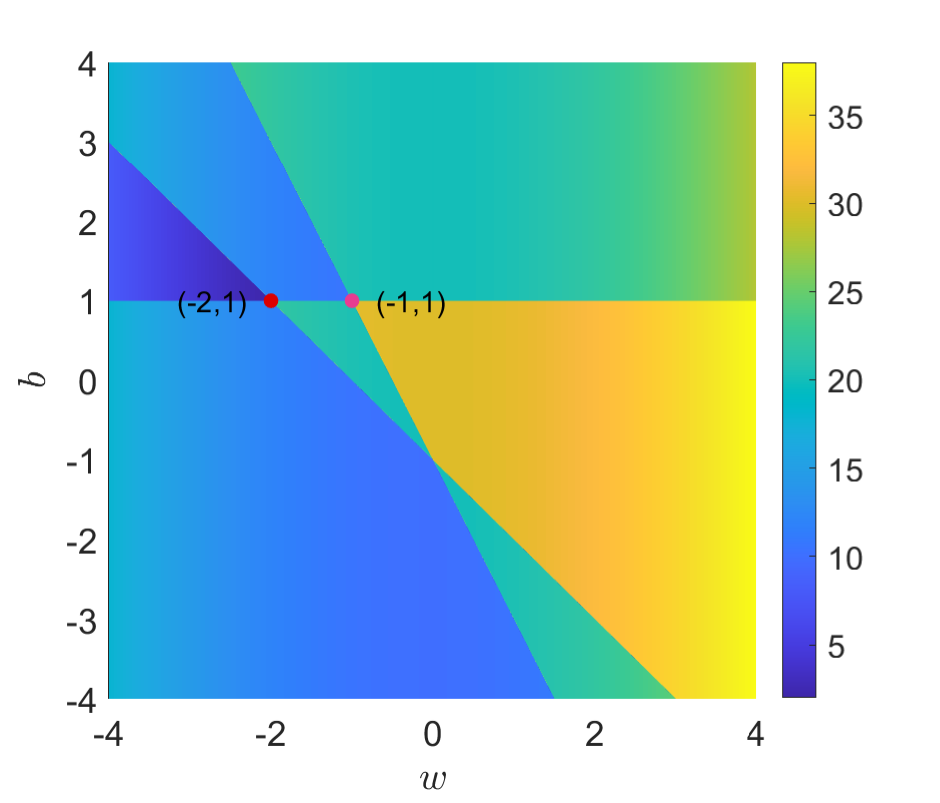}
			\end{minipage}}	
			
			\subfloat{
				\begin{minipage}[t]{0.5\linewidth}
					\centering
					\includegraphics[width=2.4in]{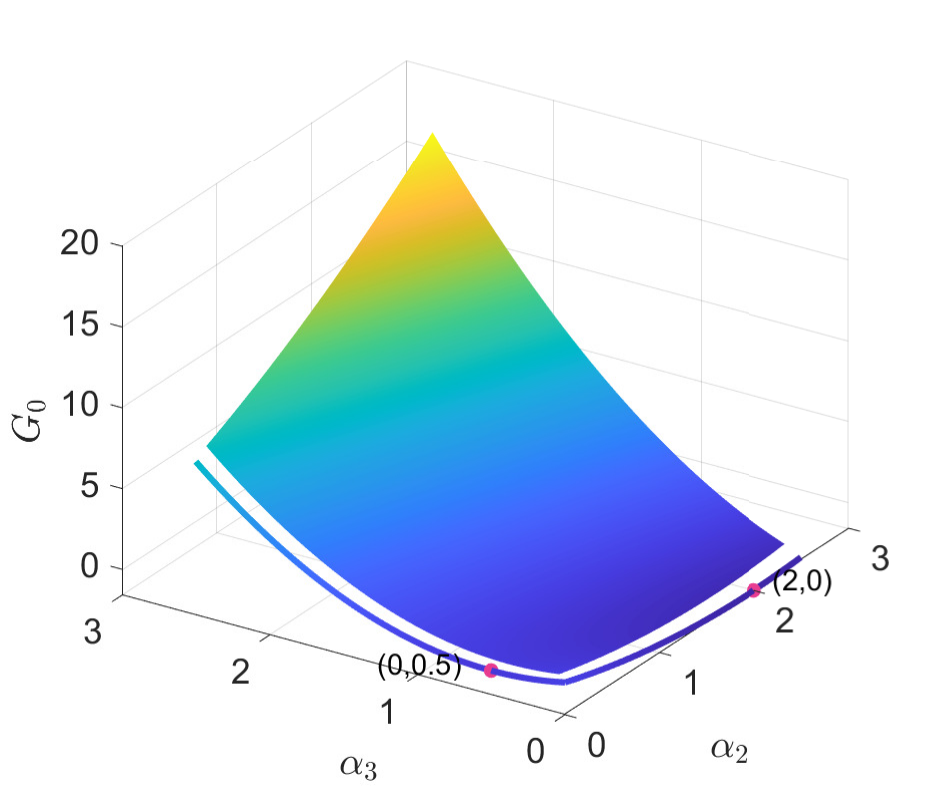}
				\end{minipage}
				\begin{minipage}[t]{0.5\linewidth}
					\centering
					\includegraphics[width=2.4in]{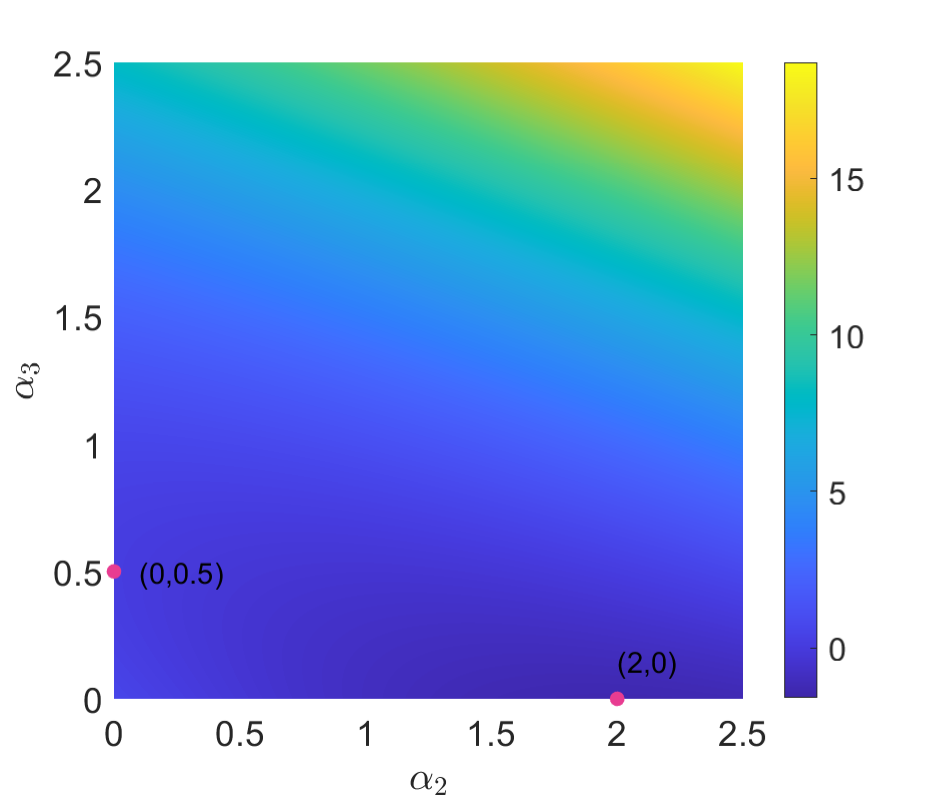}
				\end{minipage}
			}
			\caption{{Surface and contour maps of objective functions of \eqref{P01} and \eqref{D01} in Remark \ref{rmk-loc-nice-sol-interp} 
					\label{fig_minimizer}(i)}}
			{}
		\end{figure}
		We can observe that the locally nice solutions are global minimizers on their linearly constrained pieces. Moreover, each locally nice pair satisfies the linear representation and has equivalent smooth term of their respective objective functions. These observations are consist with Lemmas \ref{Lemma-Characeterization-Local-Solutions}, \ref{Lemma-Dual-Solutions} and Theorem \ref{thm-strong-duality-p0}.
		
		(ii) Statistical interpretation: The observations in Fig. \ref{fig_minimizer} indicate that the locally nice solutions tend to admit smaller objective values of \eqref{P01}, thereby leading to smaller margin loss $\sum_{i = 1}^m \ell_{0/1}( 1 - y_i (\langle \bfw, \bfx_i \rangle + b) )$ and larger margin distance $1/\| \bfw \|$. According to \cite[Theorem 2, Corollary 6]{kakade2008complexity}, these two terms help to control the expectation of misclassification rate $\mathbb{E}_{(\bfx,y)\sim\D} [\ell_{0/1} (y( \langle \bfw, \bfx \rangle + b ))]$,
		and therefore improve the generalized performance of the classifier. We believe that the locally nice solutions possess further statistical properties. This requires more detailed investigation in the future.
	\end{remark} 

\section{Implications to Hinge-Loss SVM with Positive Weights} \label{Section-Implication}

This part aims to address the relevance of 0/1-loss SVM and hSVM discussed in the Introduction.
Let us first formulate the question properly. 
We want to study the dependence of the hSVM solution in \eqref{P1} on the weight vector $\bfc$.
Recall $\bfz = [\bfw; b]$. We explicitly treat $\bfc$ in the function:
\be \label{Function-F} 
\begin{aligned}
	F_1 (\bfz, \bfc) := & \frac{1}{2} \| \bfw \|^2 + \delta_{\Re^m_+}(\bfc)  + \sum_{i = 1}^m c_i \ell_h \big( 1 - y_i (\langle \bfw, \bfx_i \rangle + b) \big), \\
	\Omega (\bfc) : = &\argmin_{\bfz} F_1 (\bfz, \bfc),
\end{aligned}
\ee 
where $\delta_{\Re^m_+}(\bfc)$ is the indicator function defined in \eqref{ind-fun}. Therefore, $\Omega(\bfc)$ is the set of global solutions of \eqref{P1} if $\bfc \geq 0$, otherwise $\Omega(\bfc)$ is empty.
Suppose we have a positive sequence $\{\bfc^\nu \}_{\nu \in \mbN}$ and the associated solutions
$\bfz^\nu \in \Omega (\bfc^\nu)$. 
Our question is whether the solutions $\bfz^\nu$ converges to a point
in $\Omega(\bfc^*)$ as $\bfc^\nu$ converges to $\bfc^*$.
This question is important as $\bfc>0$ is often dynamically set and the convergence
ensure the validity of such practices.
The answer to this question requires a known property about
uniform level boundedness in variational analysis.


This property relates the functional values of a function in one variable (e.g., $\bfz$) to another variable (e.g., $\bfc$). We say $\Psi(\bfz, \bfc)$ is level bounded in $\bfz$ locally uniformly in $\bfc$ if for
every $\widetilde{\bfc} \in \Re^m$ and $\beta \in \Re$ there is
$\varepsilon>0$ such that the set
\[
L_{\varepsilon} (\bfz, \bfc) := \left\{
(\bfz, \bfc) \ | \ \Psi(\bfz, \bfc) \le \beta, \  \bfc \in \N( \widetilde{\bfc}, \varepsilon ) 
\right\}
\]
is bounded. This uniform level-boundedness property plays an important role in the following lemma, which is an adaption from \cite[Theorem 1.17]{RockWets98}.
In order to state this lemma, we recall some concepts. We say a function
$f: \Re^n \to [-\infty, \infty]$ is proper if
$f(\bfx) > -\infty$ for all $\bfx \in \Re^n$.
We say $f$ is lower semi-continuous (lsc) at $\bfx^* \in \Re^n$ if
$
\lim\inf_{\bfx^k \rightarrow \bfx^*} f(\bfx^k) \ge f(\bfx^*).
$
Furthermore, $f$ is said to be lsc if it is lsc everywhere.

\begin{lemma} \label{lem-para-min}
	Consider a function $\Psi(\bfz, \bfc): \Re^{(n+1)} \times \Re^m \to [-\infty, \infty]$,
	which is proper, lower semi-continuous (lsc), and level bounded in $\bfz$ locally uniformly in $\bfc$. 
		Denoting $\Theta(\bfc):= \argmin_{\bfz} \Psi(\bfz, \bfc)$,
	then the following assertions hold:
	
	(i) For any $\bfc$ such that $\Psi(\cdot, \bfc) \not\equiv \infty$, we have ${\Theta(\bfc)} \not= \emptyset$.
	
	(ii) For sequence $\bfc^\nu \to \wbc$ with {$\Theta(\bfc^\nu)$} and {$\Theta(\wbc)$} being nonempty, if there exists $\wbz \in {\Theta(\wbc)}$ such that $ \lim_{\nu \to \infty}\Psi(\wbz,\bfc^\nu) = \Psi(\wbz,\wbc)$, then $\{ \bfz^\nu \}_{\nu \in \mbN}$ is bounded and each of its cluster points belongs to {$\Theta(\wbc)$}.	
\end{lemma}

Another important lemma from 
\cite[Theorem 2]{burges1999uniqueness} concerns the solution uniqueness of hSVM \eqref{P1}.

\begin{lemma} \label{lem-unique-P1}
	Let $\bfz^*=[\bfw^*; b^*]$ denote a solution of \eqref{P1} with the weight vector $\bfc \in \Re^m$. Let
	\[
	u^*_i = 1 - y_i \Big( \langle \bfw^*, \bfx_i \rangle + b^* \Big) , \quad i \in [m].
	\] 
	Then,  $\bfz^*$ is the unique global solution of \eqref{P1} if and only if 
	\begin{equation} \label{unique-P1}
		\left\{ \begin{aligned}
			& \sum_{u^*_i = 0, y_i = -1} c_i + \sum_{u^*_i > 0, y_i = -1} c_i \neq \sum_{u^*_i > 0, y_i = 1} c_i \\ 
			& \sum_{u^*_i = 0, y_i = 1} c_i + \sum_{u^*_i > 0, y_i = 1} c_i \neq \sum_{u^*_i > 0, y_i = -1} c_i.
		\end{aligned} \right.
	\end{equation}
\end{lemma}

The above two lemmas are the major tool in our proof. 
We only consider a nontrivial case, which is contained in the following assumption.
Given primal variable $\bfz^* = [\bfw^*;b^*]$ with $u^*_i = 1 - y_i( \langle \bfw^*,\bfx_i \rangle + b^*)$ for $i \in [m]$, let us denote $\S^*_+:= \{ i \in [m] \ | \ u^*_i \leq 0, y_i = 1 \}$ and $\S^*_-:= \{ i \in [m] \ | \ u^*_i \leq 0, y_i = -1 \}$.

\begin{assumption} \label{asm-S-nonempty}
	Both $\S^*_+$ and $\S^*_-$ are nonempty.
\end{assumption}

From a geometric point of view, Assumption~\ref{asm-S-nonempty} means that there exist positive and negative samples in their margin half-spaces $\{ \bfx \ | \ \langle \bfw^*, \bfx \rangle + b^* \geq 1 \}$ and $\{ \bfx \ | \ \langle \bfw^*, \bfx \rangle + b^* \leq -1 \}$. This assumption is reasonable and often easily satisfied in the SVM framework.
We note that we are only interested in the local behaviour of hinge-loss SVM near a
weight vector $\bfc^*$ with a sequence $\bfc^\nu \rightarrow \bfc^*$. 
Let $F_1(\bfz, \bfc)$ be defined in \eqref{Function-F}. 
Given $\varepsilon^*>0$, we define
\be \label{Funtion-F*}
F_*(\bfz, \bfc) := F_1(\bfz, \bfc) + \delta_{\N(\bfc^*,\veps^*)}(\bfc),
\ee
where the indicator function $\delta_{\N(\bfc^*,\veps^*)}(\bfc)$ over a neighborhood
of $\bfc^*$ restricts our interest near $\bfc^*$.
The following result establishes the behaviour of $F_*$.

\begin{lemma} \label{lem-level-bounded-p1} 
	Let $(\bfz^*, \bfal^*)$ be locally nice solutions, where $\bfz^* = [\bfw^*; b^*]$ was defined as that in Theorem~\ref{thm-strong-duality-p0}. $\I_0^*$ is defined as that
	in \eqref{Index-I0} and we define the vector $\bfc^* \in \Re^m$ by
	\begin{align} \label{c*p1}
		c^*_i \left\{ \begin{aligned}
			&> \alpha^*_i, && \mbox{if} \ \ i \in \I_0^*, \\
			& = \alpha^*_i, && \mbox{if} \ \ i \not \in \I_0^*.
		\end{aligned} \right.
	\end{align}
	
	If Assumption~\ref{asm-S-nonempty} holds, then the following assertions are true:
	
	(i) There exists $\veps^* > 0$ such that $F_*(\bfz,\bfc)$ defined in
	\eqref{Funtion-F*} is proper, lsc, and level-bounded in $\bfz$ locally uniformly in $\bfc$.
	
	(ii) It holds that $\Omega(\bfc^*) = \{\bfz^*\}$.
\end{lemma}
\bp
(i) It is easy to verify that $F_*$ is proper and lsc
because $F_1(\bfz, \bfc)$ is so and $F_*(\bfz, \bfc)$ being the sum of
$F_1(\bfz, \bfc)$ and the convex indicator function $\delta_{\N(\bfc^*,\veps^*)} (\cdot)$. We next prove its uniform level boundedness. 
Recall $\bfal^*$ satisfies the conditions in \eqref{alpha-u-relation} with index sets $\I_0^* = \{ i \in [m] \;| \; u_i^* \le 0\}$ and 
$T^* =\{ i \in [m] \; | \; \alpha^*_i > 0\}$. 
Then \eqref{c*p1} implies 
\begin{align} \label{c*+0}
	c^*_i \left\{ \begin{aligned}
		&> 0, && \mbox{if} \ \ i \in \I_0^*, \\
		& = 0, && \mbox{if} \ \ i \not\in \I_0^*.
	\end{aligned} \right.
\end{align}
Let us take $\veps^* := \min_{i \in \I_0^*}(c^*_i/2) $. 
Given $\beta \in \mbR$, to show $F_*$ is level-bounded in $\bfz$ locally uniformly in $\bfc$, we just need to prove that the following set is bounded
\begin{align*}
	\mbF := \{ (\bfz, \bfc) \ | \ \bfc \in \N(\bfc^*, \veps^*) \cap \mbR^m_+, \ F_1 (\bfz, \bfc) \leq \beta \}. 
\end{align*}
It suffices to consider the case that $\mbF$ is nonempty. 
Then for any $(\bfz, \bfc) \in \mbF$, the structure of $F_*(\bfz, \bfc)$ implies
\begin{align*}
	&	\| \bfw \|^2/2 \leq \beta, \\
	&  c_i \ell_h \big( 1 - y_i (\langle \bfw, \bfx_i \rangle + b) \big) \leq \beta \ \mbox{and} \ c_i \geq \veps^*,  \ i \in \I_0^*. 
\end{align*}
Then we have $\| \bfw \| \leq \sqrt{2\beta}$, and the second inequality above leads to
the following bound for $i \in \I_0^*$:
\begin{align*}
	&\quad  c_i \max\{ 1 - y_i (\langle \bfw, \bfx_i \rangle + b), \; 0\} \le \beta \\
	\Longleftrightarrow 
	&\quad 	1 - y_i (\langle \bfw, \bfx_i \rangle + b) \le \beta/c_i \le \beta/\varepsilon^*\\
	\Longleftrightarrow
	& \quad \begin{aligned}
		y_ib \geq & 1 - y_i \langle \bfw, \bfx_i \rangle - \beta/\veps^* \\ 
		\geq & 1 - \sqrt{2\beta} \| \bfx_i \| - \beta/\veps^*.
	\end{aligned}
\end{align*}
This together with Assumption~\ref{asm-S-nonempty} leads to the boundedness of $b$. Therefore, $\mbF$ is bounded and we have finished the proof for (i). 

(ii) We have proved in Corollary~\ref{thm-sol-01-H-SVM} that
$\bfz^*$ is a global solution of hSVM \eqref{P1} with any $\bfc$ satisfying \eqref{c*1}. 
Furthermore, \eqref{c*+0} indicates that 
\[
c^*_i = 0 \ \mbox{for} \ u^*_i > 0.
\]
With Assumption~\ref{asm-S-nonempty}, we have
\begin{align*}
	& \sum_{u^*_i = 0, y_i = -1} c_i^* + \sum_{u^*_i > 0, y_i = -1} c_i^* > 0 = \sum_{u^*_i > 0, y_i = 1} c_i^* 
\end{align*}
and
\begin{align*}
	& \sum_{u^*_i = 0, y_i = 1} c_i^* + \sum_{u^*_i > 0, y_i = 1} c_i^* > 0 = \sum_{u^*_i > 0, y_i = -1} c_i^* .
\end{align*}
That is, our choice $\bfc^*$ satisfies \eqref{unique-P1}.
According to Lemma~\ref{lem-unique-P1}, 
we must have $\bfz^*$ is the unique solution in $\Omega(\bfc^*)$ and
this is the desired conclusion.
\ep

We now state the main result in this section.

\begin{theorem} \label{thm-p0-sequential-p1}
	Under the premises of Lemma \ref{lem-level-bounded-p1}, let $\{ \bfc^\nu \}_{\nu \in \mbN}$ be a positive hyperparameter sequence and $\bfz^\nu \in \Omega(\bfc^\nu)$. If $\lim_{\nu \to \infty} \bfc^\nu = \bfc^*$, then we have $\lim_{\nu \to \infty} \bfz^\nu = \bfz^*$.
\end{theorem}
\bp
Under the premises of Lemma~\ref{lem-level-bounded-p1}, $F_*$ is proper, lsc, and level bounded in $\bfz$ locally uniformly in $\bfc$. With each $\bfc^\nu >0$ and is within the neighborhood $\N(\bfc^*, \veps^*) \cap \mbR^m_+$, hSVM defined by $F_*$ has a solution. Hence 
$\Omega(\bfc^\nu)$ and $\Omega(\bfc^*)$ are nonempty by Lemma~\ref{lem-para-min} (i). 
Then according to Lemma~\ref{lem-para-min} (ii), $\{ \bfz^\nu \}_{\nu \in \mbN}$ has a cluster point belonging to $\Omega(\bfc^*)$. Since $\Omega(\bfc^*) = \{ \bfz^* \}$ by Lemma~\ref{lem-level-bounded-p1}, we must have $\lim_{\nu \to \infty} \bfz^\nu = \bfz^*$. \ep

\begin{remark}
	Suppose $\bfc^*$ is a weight vector obtained through \eqref{c*1}.
	Then the global solution sequence $\{ \bfz^\nu \}_{\nu \in \mbN}$ from hSVM with $\bfc^\nu >0$ converges to
	a locally nice solution $\bfz^*$ as $\bfc^\nu  \rightarrow \bfc^*$.
	It raises a natural question whether a locally nice solution is any better
	than other solutions. Let us denote the objective  function value of \eqref{P01} as $F_{0/1}$. We can infer that when $\bfz^\nu$ is sufficiently close to $\bfz^*$, it holds that $F_{0/1}(\bfz^\nu) \geq F_{0/1}(\bfz^*)$ because $\bfz^*$ is a local solution of \eqref{P01}. Noting that $F_{0/1}$ is a weighted sum of a quadratic regularizer and margin loss, we can also expect that the solution $\bfz^*$ corresponds to better performance on these two terms. These conclusions will be verified by experiments in the Section \ref{Section-Numerical}.
	
\end{remark}
	\section{Implications to Ramp Loss SVM} \label{Section-ramp}
	
	As the ramp loss is closely related to the 0/1 loss, this section is devoted to further exploring the relationship between the local minimizers of \eqref{Pr} and \eqref{P01}. The following lemma relates the two loss functions and plays a 
	pivotal role in our reformulation of rSVM.
	
	\begin{lemma} \label{lem-property-lr}
		Given $\gamma>0$, we have
		\begin{align} \label{lr-envelope}
			&\ell_\gamma(t) = \min_{s \in \mbR} \left\{ |s - t| + \gamma \ell_{0/1}(s) \right\}. 
		\end{align}
		Denoting $\T_\gamma(t)$ as the solution set of \eqref{lr-envelope},
		it holds that
		\begin{align} \label{lr_prox}
			\T_\gamma(t) = \left\{ \begin{aligned}
				& t, && \mbox{if}~ t < 0~\mbox{or}~ t > \gamma \\
				& \{ 0, \gamma \}, && \mbox{if}~ t = \gamma, \\
				& 0, && \mbox{if}~ 0 \leq t < \gamma.
			\end{aligned} \right.
		\end{align}	
	\end{lemma}
	\bp
	Denoting $\psi(s):= |s - t| + \gamma \ell_{0/1}(s)$, we can compute
	\begin{align*}
		& p_1 : = \argmin_{s \geq 0} \psi(s) = \max\{ t,0 \},\\
		& q_1 := \min_{s \geq 0} \psi(s) = |\max\{t,0\} - t| + \gamma = -\min\{ t, 0\} + \gamma, \\
		& p_2 := \argmin_{s \leq 0} \psi(s) = \min\{ t,0 \}, \\
		& q_2 := \min_{s \leq 0} \psi(s) = | \min\{ t,0 \} - t | = \max\{ t, 0 \}.
	\end{align*}
	Then to calculate $\min_s \psi(s)$ and the associated solutions, we just need compare $q_1$ and $q_2$.
	
	\underline{If $q_1 < q_2$, which means $t > \gamma$}
	due to $\gamma < \max\{t, 0\} + \min\{t, 0\}=t$, then $\argmin_s \psi(s) = p_1 = t$ and $\min_s \psi(s) = q_1 = \gamma$. 
	
	\underline{If $q_1 = q_2$, and hence $t = \gamma$}, then $\argmin_s \psi(s) = \{p_1, p_2\} = \{ 0, \gamma \}$ and $\min_s \psi(s) = \gamma$.
	
	\underline{If $q_1 > q_2$, which means $t < \gamma$}, then $\argmin_s \psi(s) = p_2 $ and $\min_s \psi(s) = q_2 = \max\{ t, 0 \}$. 
	
	The above three cases imply the conclusion of this lemma.
	\ep
	
	This lemma implies that the ramp loss is an $\ell_1$-norm envelope of 0/1-loss. As a result, 
	Problem \eqref{Pr} can be equivalently reformulated as 
	the following problem with an extra variable $\bfv \in \Re^m$ (the optimal objectives are equal):
	\begin{align} \label{penalty-l01}
		\min_{\bfz, \bfv}~ &\frac{1}{2} \|\bfw\|^2 + \rho \sum_{i = 1}^m |1 - y_i (\langle\bfw,\bfx_i \rangle + b) - v_i| + \rho \gamma \sum_{i = 1}^m  \l01(v_i) .  
	\end{align}
	Moreover, if $\bfz^*$ is a local minimizer of \eqref{Pr}, then $( \bfz^*, \bfxi^* )$ must be a local minimizer of \eqref{penalty-l01}, where for $i \in [m],$
	\begin{align} \label{def-xi}
		\xi_i(\bfz):= \T_r( 1 - y_i ( \langle \bfw, \bfx_i \rangle + b) )~ \mbox{and}~ \xi^*_i := \xi_i(\bfz^*).
	\end{align}
	Notably, \eqref{penalty-l01} turns out to be the $\ell_1$-norm penalty problem of \eqref{P01}. It serves as an intermediate model in establishing the relationship between \eqref{P01} and \eqref{Pr}. Noticing that \eqref{penalty-l01} has the similar structure to that of \eqref{P01}, we can also use a convex program (just like \eqref{hm-SVM-I}) to characterize the local minimizers of \eqref{penalty-l01}.
	
	Given $\I \subseteq [m]$, let us consider the convex optimization
	\begin{align} 
		\min_{\bfz, \bfv}~ & \Phi(\bfz, \bfv) := \frac{1}{2} \|\bfw\|^2 + \rho \sum_{i = 1}^m |1 - y_i (\langle\bfw,\bfx_i \rangle + b) - v_i| \notag \\
		s.t.~ &v_i \leq 0,~~ i \in \I. \label{penalty-l01-I}
	\end{align}
	The global minimizer of this problem can be characterized by the following KKT system:
	\begin{align} \label{KKT-penalty-l01-I} 
		\left\{ \begin{aligned}
			& \bfw= \sum_{i=1}^m \alpha_i y_i  \bfx_i, \\
			& \alpha_i \in \partial \rho | 1 - y_i ( \langle \bfw, \bfx_i \rangle + b) - v_i |,~i \in [m] \\
			& v_i \le 0 , \ \alpha_i \ge 0, \ \alpha_i v_i = 0 , \ i \in \I, \\
			& \alpha_i = 0, \ \ i \not\in \I, \quad \langle \bfy, \bfal \rangle = 0,
		\end{aligned} \right.
	\end{align}
	To proceed, given $\bfv$ and $\bfv^*$, let us denote the following sets:
	\begin{align*}
		\left\{ \begin{aligned}
			&\Gamma_0(\bfv): = \{ i \in [m] ~|~ v_i \leq 0 \},~ \Gamma_+(\bfv): = \{ i \in [m] ~|~ v_i > 0 \}, \\
			&\Gamma^*_0: = \Gamma_0(\bfv^*),~ \Gamma^*_+: = \Gamma_+(\bfv^*), \\
			& \F^*_0 := \{ (\bfz, \bfv) ~|~ v_i \leq 0, ~ i \in \Ga_0^* \}.
		\end{aligned} \right.
	\end{align*}
	Following a similar procedure to Lemma \ref{Lemma-Characeterization-Local-Solutions}, we can obtain the following lemma. 
	\begin{lemma} \label{local-min-penalty-l01}
		The point $(\bfz^*, \bfv^*)$ is a local minimizer of \eqref{penalty-l01} if and only if $(\bfz^*, \bfv^*)$ is a global minimizer of \eqref{penalty-l01-I} with $\I = \Ga_0^*$.
	\end{lemma}
	\bp
	If $(\bfz^*,\bfv^*)$ is a local minimizer of \eqref{penalty-l01}, then there exists $\tau > 0$ such that for any $(\bfz, \bfv) \in \N( (\bfz^*,\bfv^*), \tau )$, we have
	\begin{align*}
		\Phi(\bfx, \bfv) +  \rho \gamma \underbrace{\sum_{i = 1}^m  \l01(v_i)}_{|\Gamma_+(\bfv)|} \geq \Phi(\bfx^*, \bfv^*) +  \rho \gamma \underbrace{\sum_{i = 1}^m  \l01(v^*_i)}_{|\Gamma_+^*|}
	\end{align*}
	Then considering $(\bfz, \bfv) \in \N( (\bfz^*,\bfv^*), \tau ) \cap \F^*_0$, it holds that $|\Ga_+(\bfv)| \leq |\Ga_+^*|$. This together with the above inequality implies $\Phi(\bfx, \bfv) \geq \Phi(\bfx^*, \bfv^*)$, which means that $(\bfz^*, \bfv^*)$ is a local minimizer of \eqref{penalty-l01-I} with $\I = \Gamma_0^*$. Since this problem is convex, $(\bfz^*, \bfv^*)$ is also its global minimizer.
	
	Conversely, if $(\bfz^*, \bfv^*)$ is a global minimizer of \eqref{penalty-l01-I} with $\I = \Gamma_0^*$, then we have
	\begin{align} \label{global-minimizer-Phi}
		\Phi(\bfx, \bfv) \geq \Phi( \bfx^*, \bfv^* ), ~~ \forall~(\bfx, \bfv) \in \F^*_0.
	\end{align}
	Let us take a sufficiently small radius $\tau_0 > 0$ such that for any $(\bfz, \bfv) \in \N( (\bfz^*,\bfv^*), \tau_0 )$, we have
	\begin{align} \label{continuity-Phi}
		\Phi(\bfx, \bfv) \geq \Phi( \bfx^*, \bfv^* ) - \rho \gamma/2 ~\mbox{and}~ \Gamma_+(\bfv) \supseteq \Ga_+^*,
	\end{align}
	where the first formula follows from the continuity of $\Phi$. Since \eqref{global-minimizer-Phi} holds in $\F^*_0$ whereas \eqref{continuity-Phi} holds in $\N( (\bfz^*,\bfv^*), \tau_0 )$, we consider the following two cases to show $(\bfz^*, \bfv^*)$ is a local minimizer of \eqref{penalty-l01}.
	
	\underline{If $(\bfz, \bfv) \in \N( (\bfz^*,\bfv^*), \tau_0 ) \cap \F^*_0$}, then \eqref{global-minimizer-Phi} and the second formula in \eqref{continuity-Phi} implies 
	\begin{align*}
		\Phi(\bfx, \bfv) +  \rho \gamma \sum_{i = 1}^m  \l01(v_i) \geq \Phi(\bfx^*, \bfv^*) +  \rho \gamma \sum_{i = 1}^m  \l01(v^*_i)
	\end{align*}
	\underline{If $(\bfz, \bfv) \in \N( (\bfz^*,\bfv^*), \tau_0 ) \cap \overline{\F}^*_0$}, then there exists $i_0 \in \Ga_0^*$ such that $v_{i_0} >0$, and hence $| \Ga_0(\bfv) | \geq | \Ga_0^* | + 1$. Combining this with the first formula in \eqref{continuity-Phi}, we can derive 
	\begin{align*}
		\Phi(\bfx, \bfv) +  \rho \gamma \sum_{i = 1}^m  \l01(v_i) \geq \Phi(\bfx^*, \bfv^*) +  \rho \gamma \sum_{i = 1}^m  \l01(v^*_i) + \frac{\rho \gamma}{2}
	\end{align*} 
	Overall, the above two cases imply the desired conclusion.
	\ep

	Given a local minimizer $\bfz^*$ of \eqref{P01}, Lemma \ref{Lemma-Characeterization-Local-Solutions} indicates that there must exist multiplier $\bfal^*$ such that the following KKT condition holds:
	\begin{align} \label{KKT-PI} 
		\left\{ \begin{aligned}
			& \bfw^*= \sum_{i=1}^m \alpha_i^* y_i  \bfx_i, \\
			& u^*_i = 1 - y_i ( \langle \bfw^*, \bfx_i \rangle + b^*), \ i \in [m], \\
			& u^*_i \le 0 , \ \alpha_i^* \ge 0, \ \alpha^*_i u^*_i = 0 , \ i \in \I_0^* \\
			& \alpha^*_i = 0, \ \ i \not\in \I_0^*, \quad \langle \bfy, \bfal^* \rangle = 0.
		\end{aligned} \right.
	\end{align}
	Next we will show that $\bfz^*$ is also a local minimizer of \eqref{Pr} when $\rho$ and $\gamma$ are properly selected according to $\bfal^*$ and $\bfu^*$ respectively.
	\begin{theorem} \label{theorem-P01-Pr}
		Let $\bfz^*$ be a local minimizer of \eqref{P01} and $\bfal^*$ be a multiplier satisfying \eqref{KKT-PI}. If we take
		\begin{align} \label{rho-r}
			\rho > \max_{i \in [m]} \alpha^*_i ~\mbox{and}~ 0 < \gamma < \min\{ u^*_i~|~u^*_i > 0 \},
		\end{align}
		then $\bfz^*$ is a local minimizer of \eqref{Pr}.
	\end{theorem}
	\bp
	There are two steps in this proof. 
	
	\uline{Step 1: If $\rho > \max_{i \in [m]} \alpha^*_i$, then $(\bfz^*, \bfu^*)$ is a local minimizer of \eqref{penalty-l01}.} 
	To prove this assertion, we just need show that $(\bfz^*,\bfu^*, \bfal^*)$ satisfy \eqref{KKT-penalty-l01-I} with $\I = \Gamma_0(\bfu^*)$ by Lemma \ref{local-min-penalty-l01}. We can observe $\Gamma_0(\bfu^*) = \I^*_0$ by their definitions. 
	Then by comparing \eqref{KKT-PI} and \eqref{KKT-penalty-l01-I}, it suffices to prove $\alpha^*_i \in \partial \rho | 1 - y_i ( \langle \bfw^*, \bfx_i \rangle + b^*) - u^*_i |$. This must be true due to $\partial\rho | 1 - y_i ( \langle \bfw^*, \bfx_i \rangle + b^*) - u^*_i | = [- \rho, \rho]$ and $\rho > \max_{i \in [m]} \alpha^*_i$. 
	
	\uline{Step 2: If $0 < \gamma < \min\{ u^*_i~|~u^*_i > 0 \}$ further holds, then $(\bfz^*, \bfu^*)$ is a local minimizer of \eqref{Pr}.} The claim of Step 1 means there exists $\tau_1 > 0$ such that for any $(\bfz, \bfv) \in \N( (\bfz^*,\bfu^*), \tau_1 )$, we have 
	\begin{align}
		&\frac{1}{2} \|\bfw\|^2 + \rho \sum_{i = 1}^m \big( |1 - y_i (\langle\bfw,\bfx_i \rangle + b) - v_i| + \gamma  \l01(v_i) \big) \notag  \\
		\geq& \frac{1}{2} \|\bfw^*\|^2 + \rho \sum_{i = 1}^m\big( |1 - y_i (\langle\bfw^*,\bfx_i \rangle + b^*) - u^*_i| + \gamma  \l01(u^*_i) \big) \notag \\
		\geq& \frac{1}{2} \|\bfw^*\|^2 + \rho \sum_{i = 1}^m \ell_\gamma \big( 1 - y_i (\langle\bfw^*,\bfx_i \rangle + b^*) \big), \label{loc-min-penalty-l01}
	\end{align}
	where the last inequality follows from \eqref{lr-envelope}. When taking $0 < \gamma < \min\{ u^*_i~|~u^*_i > 0 \}$, it follows from \eqref{lr_prox} that $\T_{\gamma}(u^*_i) = u^*_i$, and $\T_{\gamma}(\cdot)$ is single-valued and continuous around $u^*_i$ for $i \in [m]$. 
	Recalling the definition of $\bfxi(\bfz)$ in \eqref{def-xi}, there exists $\tau_2 > 0$ such that for any $\bfz \in \N(\bfz^*, \tau_2)$,
	\begin{align*}
		| \xi_i(\bfz) - \T(u^*_i) | = | \xi_i(\bfz) - u_i^* | < \tau_1/(2\sqrt{m}),~ i\in [m].
	\end{align*}
	This implies that if $\| \bfz - \bfz^* \| < \min\{ \tau_2, \tau_1 \}/2$, then $(\bfz, \bfxi(\bfz)) \in \N((\bfz^*, \bfu^*), \tau_1) $. Finally, by substituting $(\bfz, \bfxi(\bfz))$ into the left-hand side of \eqref{loc-min-penalty-l01} and then using Lemma \ref{lem-property-lr}, we can prove that for any $\bfz \in \N( \bfz^*, \min\{ \tau_2, \tau_1 \}/2 )$, it holds that
	\begin{align*}
		&\frac{1}{2} \|\bfw\|^2 + \rho \sum_{i = 1}^m \ell_\gamma \big( 1 - y_i (\langle\bfw,\bfx_i \rangle + b) \big) \\
		\geq &\frac{1}{2} \|\bfw^*\|^2 + \rho \sum_{i = 1}^m \ell_\gamma \big( 1 - y_i (\langle\bfw^*,\bfx_i \rangle + b^*) \big) 
	\end{align*}
	This is exactly the desired conclusion. \ep
	
	{Geometrically, the ramp loss approximates 0/1 loss more closely when $\rho$ is large and $\gamma$ is small. This provides an intuitive interpretation for Theorem \ref{theorem-P01-Pr}. At the end of this section, let us discuss the robustness and support vectors of \eqref{P01} and \eqref{Pr}.
		\begin{remark}
			(i) It has been widely recognized that both 0/1 loss and ramp loss increase the robustness of SVM to outliers (see e.g. \cite{wu2007robust,brooks2011support,nguyen2013algorithms,wang2021support,zhou2021global}). Particularly, in \cite[Example 5.1]{zhou2021quadratic} and \cite[Exaxmple 5.2]{wang2021support}, 0/1-loss SVM yields a better decision hyperplane and higher accuracy than that of ramp-loss SVM on several simulated data with outliers. More comprehensive evaluations about robustness of the two SVMs can be found in \cite{brooks2011support}. Since our work focuses on the relationship between the solutions of the SVMs, we refer the reader to these references for further details. 	
			%
			
			(ii) If $\bfz^*$ is a local minimizer of 0/1-loss SVM \eqref{P01}, then there exists multiplier $\bfal^*$ satisfies KKT condition \eqref{KKT-PI}. Particularly, the third and fourth lines imply that $u^*_i = 0$ if $\alpha^*_i \neq 0$. This means that the associated support vectors belong to $\Omega_1:=\{ \bfx_i ~|~ y_i( \langle \bfw^*, \bfx_i \rangle + b^*  ) = 1  \}$, lying on the margin hyperplane. If $\bfz^*$ is a local minimizer of ramp-loss SVM \eqref{Pr}, then $(\bfz^*, \bfxi^*)$ is a local minimizer of \eqref{penalty-l01}. Then it follows from Lemma \ref{local-min-penalty-l01} that $(\bfz^*, \bfxi^*)$ and the associated multiplier $\bfal^*$ satisfy \eqref{KKT-penalty-l01-I} with $\I = \Ga_0(\bfxi^*)$. The third and fourth lines of \eqref{KKT-penalty-l01-I} indicate that $\xi^*_i = 0$ if $\alpha^*_i \neq 0$. By the definition of $\bfxi^*$ (see \eqref{def-xi}), we know that all the support vectors of $\bfz^*$ belong to the set $\Omega_2:=\{ \bfx_i ~|~ 1 - \gamma \leq y_i( \langle \bfw^*, \bfx_i \rangle + b^*  ) \leq 1 \}$. On comparison, \eqref{P01} tends to admit less support vectors than that of \eqref{Pr}. 
		\end{remark}
	}
	
	{
		\begin{remark} \label{rem-support-vector}
			%
			As there are multiple local minimizers of \eqref{P01}, a more in-depth issue concerns defining a stronger notion of support vectors. Let us consider an example with the samples $\bfx_i \in \{ (0,0), (0.5,0.5), (0,1), (1,0) \}$ and labels $y_i \in \{ 1, 1, -1, -1 \}$. By enumerating index $\I \subseteq \{ 1,2,3,4 \}$ and solving \eqref{hm-SVM-I}, we can identify all the locally nice pairs of \eqref{P01} with nonempty support vectors.
			{	
				\begin{table}[H]
					\caption{Locally nice solutions in Remark \ref{rem-support-vector}}
					\centering
					{
						\begin{tabular}{c|c|c}
							\hline
							$\bfz^* = [\bfw^*;b^*]$            & $\bfal^*$       & Support Vectors         \\ \hline
							$(0;-2;1)$  & $(2;0;2;0)$ & $\bfx_1, \bfx_3$        \\
							$(-2;0;1)$  & $(2;0;0;2)$ & $\bfx_1, \bfx_4$        \\
							$(-2;-2;1)$ & $(4;0;2;2)$ & $\bfx_1, \bfx_3,\bfx_4$ \\
							$(-2;2;1)$  & $(0;4;0;4)$ & $\bfx_2, \bfx_4$        \\
							$(2;-2;1)$  & $(0;4;4;0)$ & $\bfx_2, \bfx_3$        \\ \hline
					\end{tabular}}
			\end{table}}
			Then we can compute the frequency that $\bfx_i$ for $i= 1,\cdots,4$ being a support vector are 0.6, 0.4, 0.6 and 0.6 respectively. Therefore, $\bfx_1$, $\bfx_3$, and $\bfx_4$ are more possible to be support vectors and we call them ``the most relevant samples". In this example, these points are exactly the vertexes of the convex hull of the data points. We believe that the most relevant samples possess additional geometric or statistical properties, but a rigorous characterization remains to be developed. The foremost challenge is how to efficiently identify the most relevant samples when $m$ is large. 
			Enumeration of all local solutions through a mixed integer programming
			\cite{tillmann2024cardinality} is less efficient even 
			when $m$ is moderate.
			We leave this topic to our future work.
		\end{remark}
	}

	\section{Numerical Experiments} \label{Section-Numerical}
	
	The purpose of this section is to illustrate that a local solution of 0/1-loss SVM can achieve better classification performance than global solutions of hSVM and local solutions of rSVM with certain range of hyperparameters. All the experiments in this section are conducted on Matlab 2022a by a laptop
	with 32GB memory and Intel CORE i7 2.6 GHz CPU. The details of datasets for experiments are summarized in Tab. \ref{tab-dataset}
	
	\begin{table}[H]
		\caption{{Real binary classification datasets}} 
		\label{tab-dataset}
		\centering
		{\begin{tabular}{cccc}
				\hline
				Dataset              & Source & $m$  & $n$ \\ \hline
				\texttt{a1a}         & libsvm$\tablefootnote{https://www.csie.ntu.edu.tw/~cjlin/libsvm/\label{libsvm}}$ & 1605 & 123 \\
				\texttt{german}     & libsvm & 1000 & 24  \\
				\texttt{heart\_scale} & libsvm & 270  & 13  \\
				\texttt{madeline}    & openml$\tablefootnote{https://openml.org/\label{openml}}$  & 3140 & 260 \\
				\texttt{credit\_approval}         & openml & 1000 & 50   \\
				\texttt{sonar}                   & openml & 208  & 60   \\
				\texttt{spect}                   & openml & 267  & 22   \\
				\texttt{wisconsin\_breast\_cancer} & openml & 699  & 10 \\  \hline
		\end{tabular}}
	\end{table}
	
	{Now let us describe the procedure of experiments. In \cite{wang2021support}, $L_{0/1}$ADMM is developed for solving 0/1-loss SVM. The augmented Lagrangian function of this problem is defined as
		\begin{align*}
			\mathcal{L}_\sigma( \bfz, \bfv, \bfal ) := \frac{1}{2} \| \bfw \|^2 + \langle \bfal, A\bfz + \bfone - \bfv \rangle + \frac{\sigma}{2} \| A\bfz + \bfone - \bfv \|^2 + \lambda \sum_{i=1}^{m} \ell_{0/1}(v_i),
		\end{align*}
		where $\sigma > 0$ and matrix $A$ has been defined in \eqref{data-A}. By alternatively optimizing the augment Lagrangian function with respect to variables $\bfz$, $\bfv$, and $\bfal$, $L_{0/1}$ADMM generates a sequence $\{ (\bfz^k, \bfv^k, \bfal^k) \}_{k \in \mbN}$. Particularly, the limiting point of primal-dual sequence $\{ (\bfz^k, \bfal^k) \}_{k \in \mbN}$ is a pair of locally nice solution $(\bfz^*, \bfal^*)$ under suitable conditions. In our experiments, all the algorithmic parameters of $L_{0/1}$ADMM are set in their default values.}

	{To construct hyperparameter vector $\bfc$ satisfying \eqref{c*p1}, we take
		\begin{align*}
			c_i = \left\{ \begin{aligned}
				& \alpha^*_i + \eta, && \mbox{if} \ \ i \in \I^*_0, \\
				&\alpha^*_i, && \mbox{if} \ \ i \notin \I^*_0,
			\end{aligned} \right.
		\end{align*}
		where $\eta = 10^{-15}$. After that, we select a positive hyperparameter sequence $\bfc^\nu = \max\{ \bfc + \nu \textbf{1}, \eta \}$ and $\nu \in [-10^{4}, 10^4]$. Finally, for each $\bfc^\nu$, a global solution $\bfz^\nu_h$ of \eqref{P1} is computed by the solver LIBSVM \cite{chang2011libsvm}. We take $\eta$ as a small positive quantity for the following reasons. Firstly, in this setting, $\bfc^\nu$ is strictly positive, which satisfies the requirement for the input variable of the solver LIBSVM. Secondly, $\bfc$ satisfies theoretical setting \eqref{c*p1}, and $\bfc^0$ is sufficiently close to $\bfc$ because of $\| \bfc^0 - \bfc \| \leq m/10^{15}$. This allows us to observe the performance of the
		solutions near $\bfc$.
	} 
	
	
	{Next, we take $\rho = \max_{i \in [m]} \alpha^*_i + \eta$ and $\gamma = \min\{ u^*_i~|~u^*_i > 0 \} - \eta$ to satisfy \eqref{rho-r}, where $\eta = 10^{-15}$. Then hyperparameter sequences are selected by $\rho^\nu = \max\{ \rho + \nu, \eta \}$ and $\gamma^\nu = \max\{ \gamma + \nu, \eta \}$, where $\nu \in [-10^4, 10^4]$. In this way, the hyperparameter sequences are positive. Moreover, $\rho^0$ and $\gamma^0$ are sufficiently close to $\rho$ and $\gamma$ respectively. For each $\rho^\nu$ and $\gamma^\nu$, we solve \eqref{Pr} by a concave-convex procedure (CCCP, see \cite{collobert2006trading}). Since rSVM is nonconvex optimization, we can only ensure to find a local minimizer $\bfz^\nu_r$. }
	
\begin{figure*}[htb]
		\subfloat{
			\begin{minipage}[t]{0.25\linewidth}
				\centering
				\includegraphics[width=1.5in]{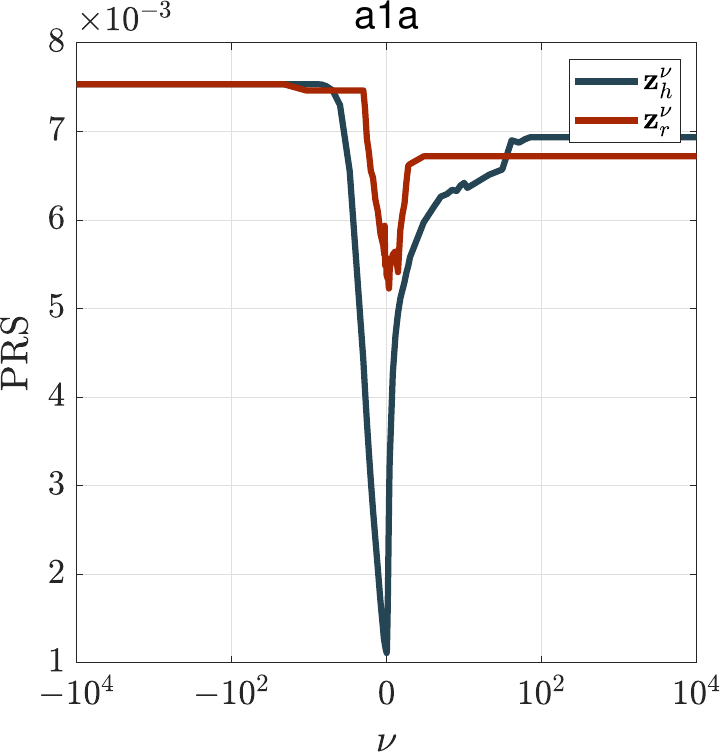}
			\end{minipage}%
			\begin{minipage}[t]{0.25\linewidth}
				\centering
				\includegraphics[width=1.6in]{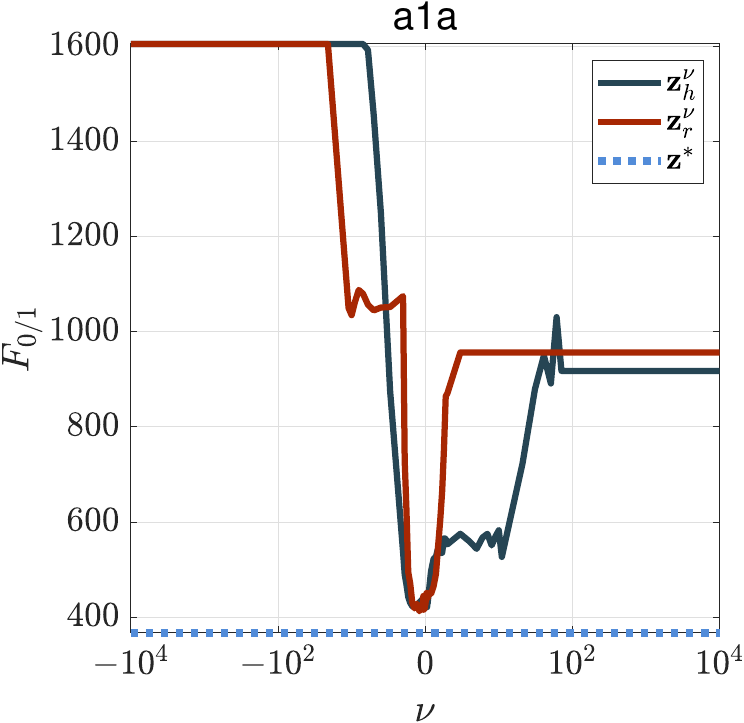}
			\end{minipage}%
			\begin{minipage}[t]{0.25\linewidth}
				\centering
				\includegraphics[width=1.55in]{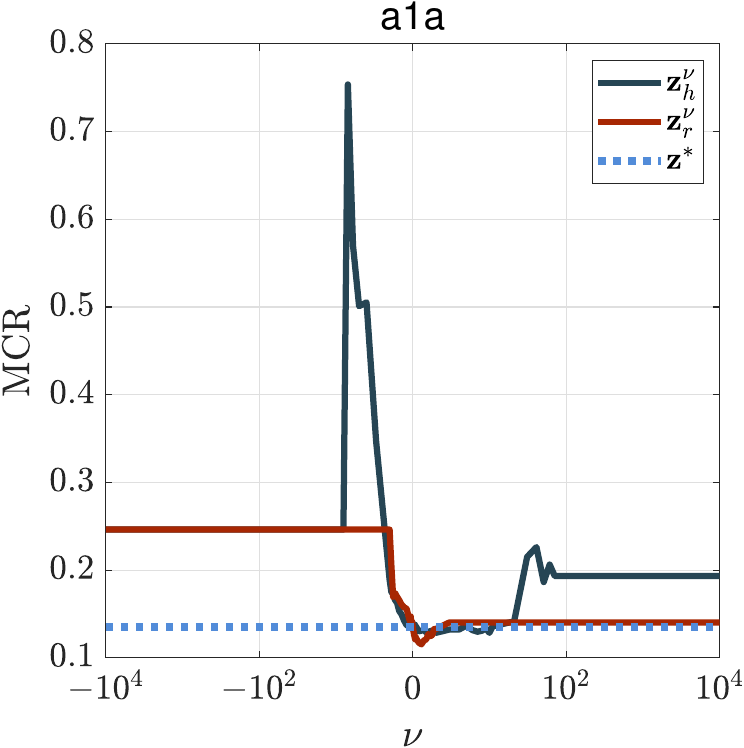}
			\end{minipage}
			\begin{minipage}[t]{0.25\linewidth}
				\centering
				\includegraphics[width=1.6in]{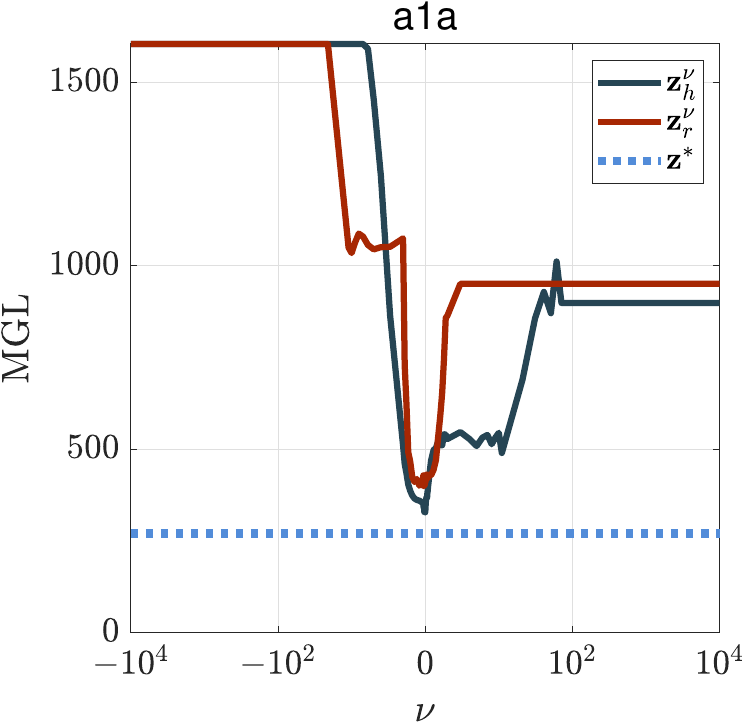}
		\end{minipage}}
		\caption{{Numerical results on dataset \texttt{a1a}.}} \label{fig-a1a}
		{}
	\end{figure*}

\begin{figure*}[htb]
		\subfloat{
			\begin{minipage}[t]{0.25\linewidth}
				\centering
				\includegraphics[width=1.6in]{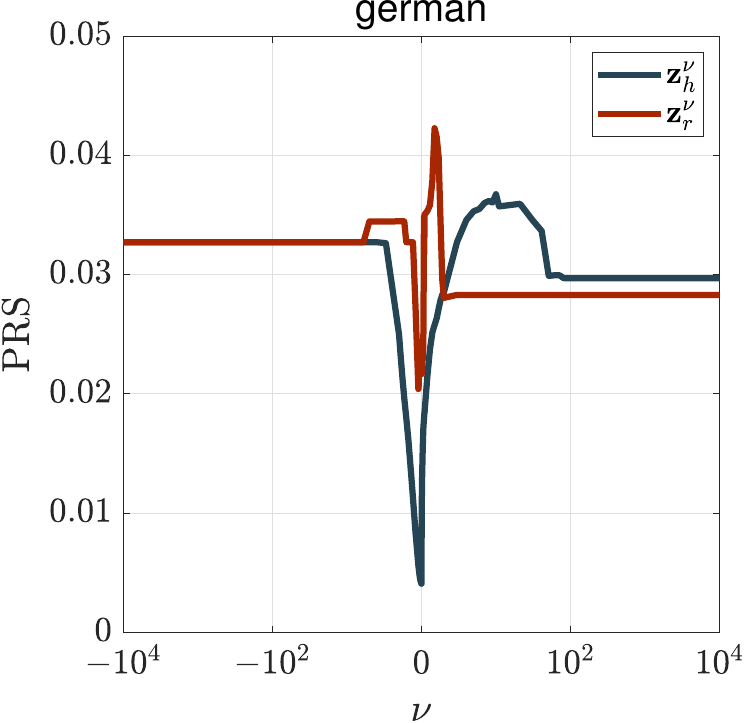}
			\end{minipage}%
			\begin{minipage}[t]{0.25\linewidth}
				\centering
				\includegraphics[width=1.6in]{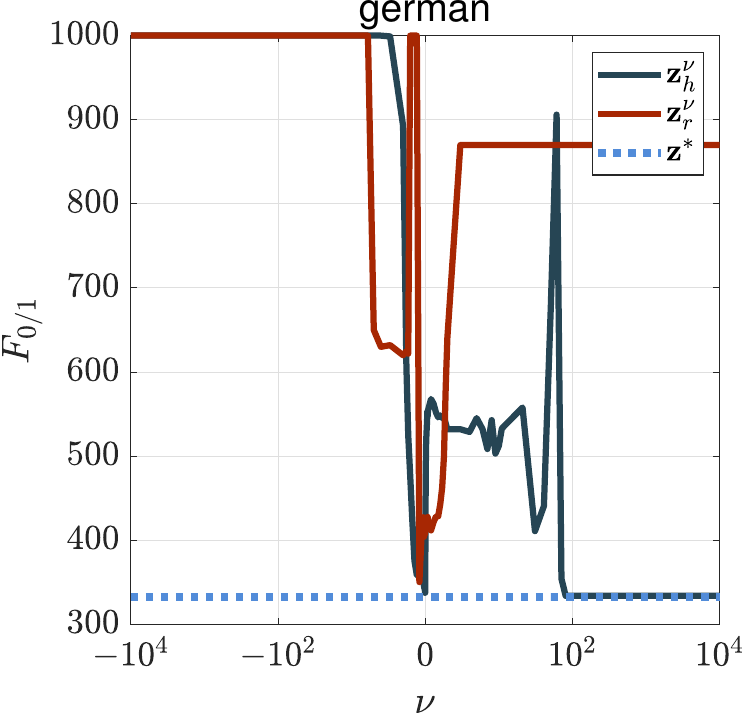}
			\end{minipage}%
			\begin{minipage}[t]{0.25\linewidth}
				\centering
				\includegraphics[width=1.53in]{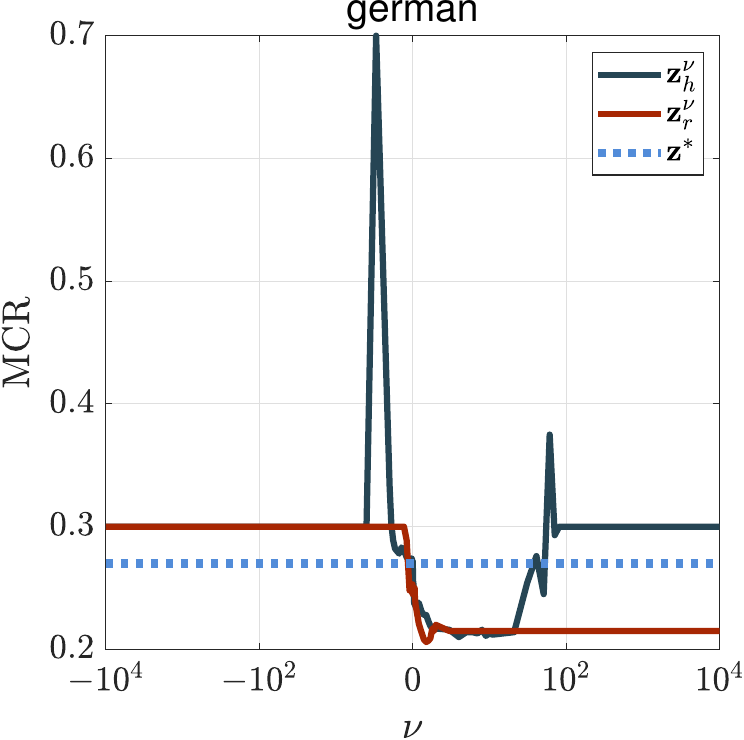}
			\end{minipage}
			\begin{minipage}[t]{0.25\linewidth}
				\centering
				\includegraphics[width=1.55in]{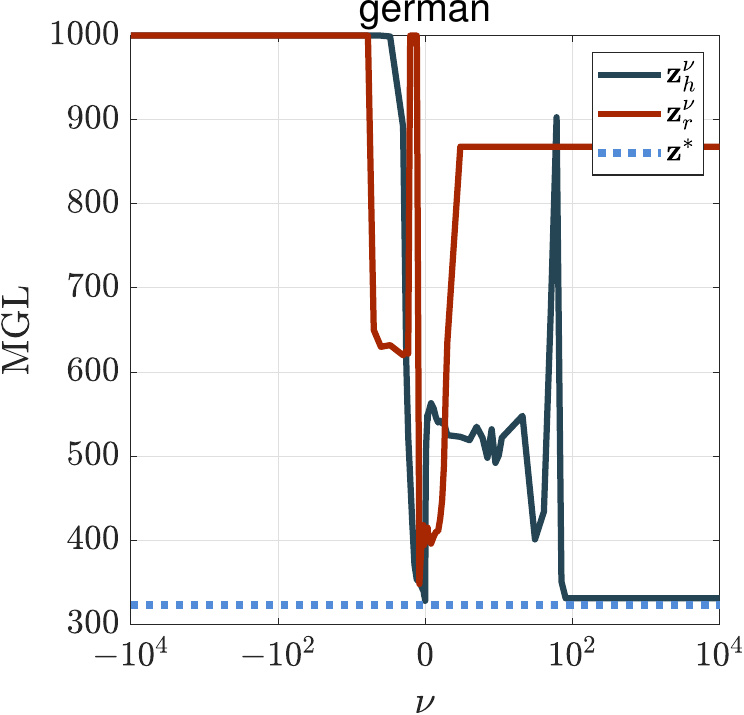}
		\end{minipage}}
		\caption{{Numerical results on dataset \texttt{german}.}} \label{fig-german}
		{}
	\end{figure*}

\begin{figure*}[htb]
		\subfloat{
			\begin{minipage}[t]{0.25\linewidth}
				\centering
				\includegraphics[width=1.6in]{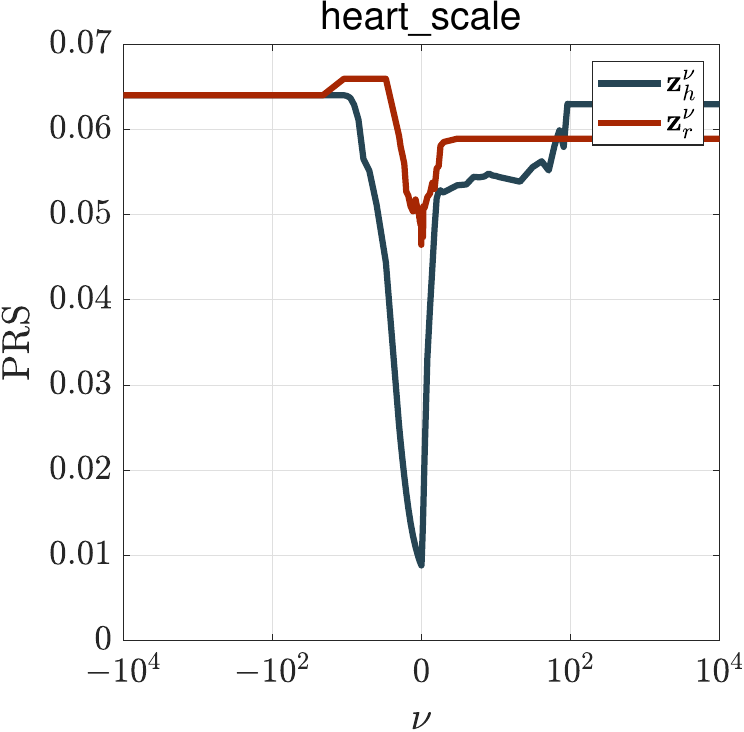}
			\end{minipage}%
			\begin{minipage}[t]{0.25\linewidth}
				\centering
				\includegraphics[width=1.6in]{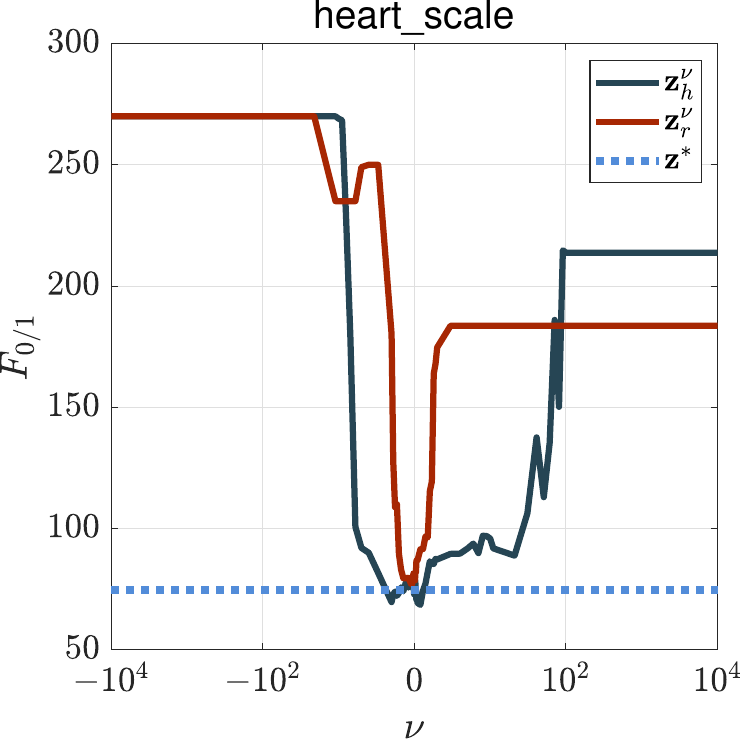}
			\end{minipage}%
			\begin{minipage}[t]{0.25\linewidth}
				\centering
				\includegraphics[width=1.6in]{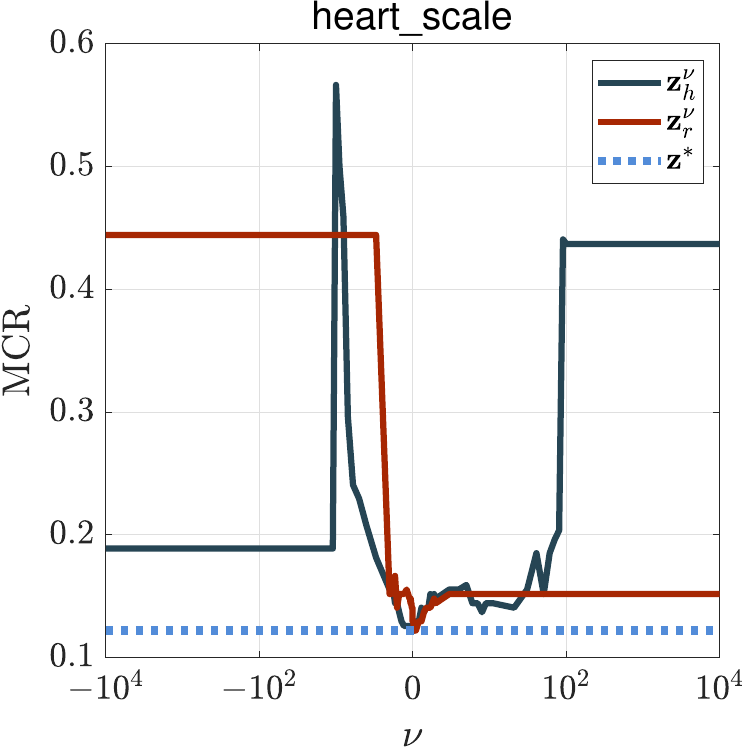}
			\end{minipage}
			\begin{minipage}[t]{0.25\linewidth}
				\centering
				\includegraphics[width=1.6in]{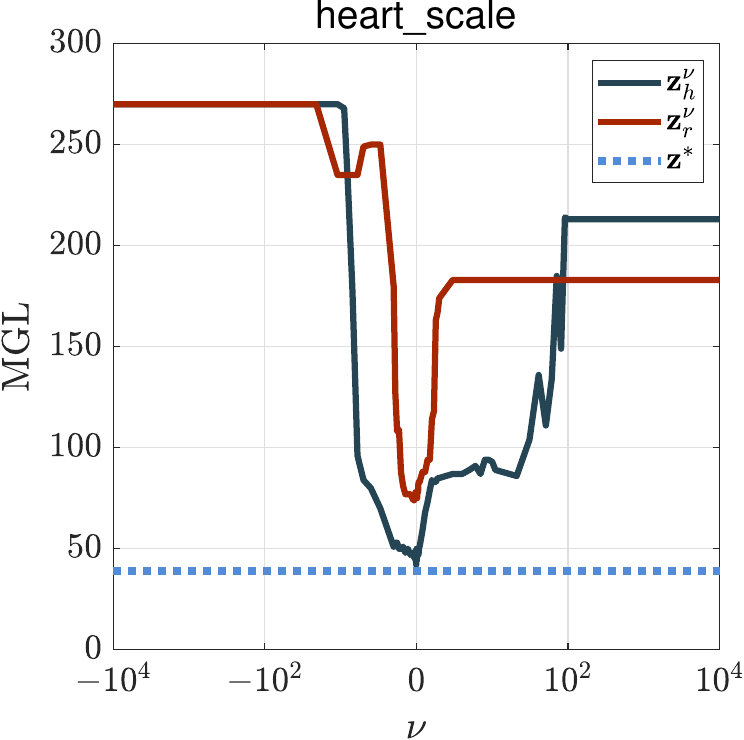}
		\end{minipage}}
		\caption{{Numerical results on dataset \texttt{heart\_scale}.}} \label{fig-heart_scale}
		{}
	\end{figure*}

\begin{figure*}[htb]
		\subfloat{
			\begin{minipage}[t]{0.25\linewidth}
				\centering
				\includegraphics[width=1.55in]{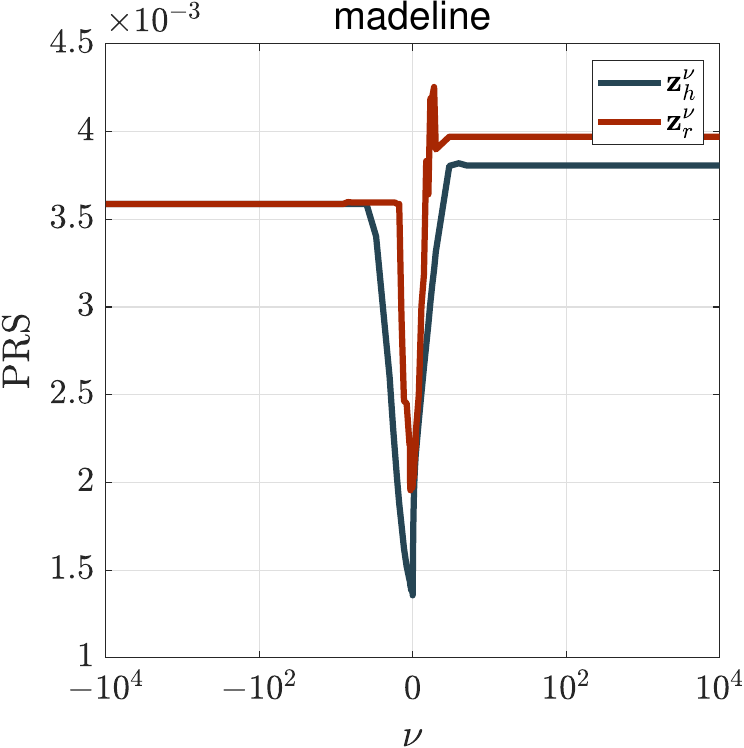}
			\end{minipage}%
			\begin{minipage}[t]{0.25\linewidth}
				\centering
				\includegraphics[width=1.6in]{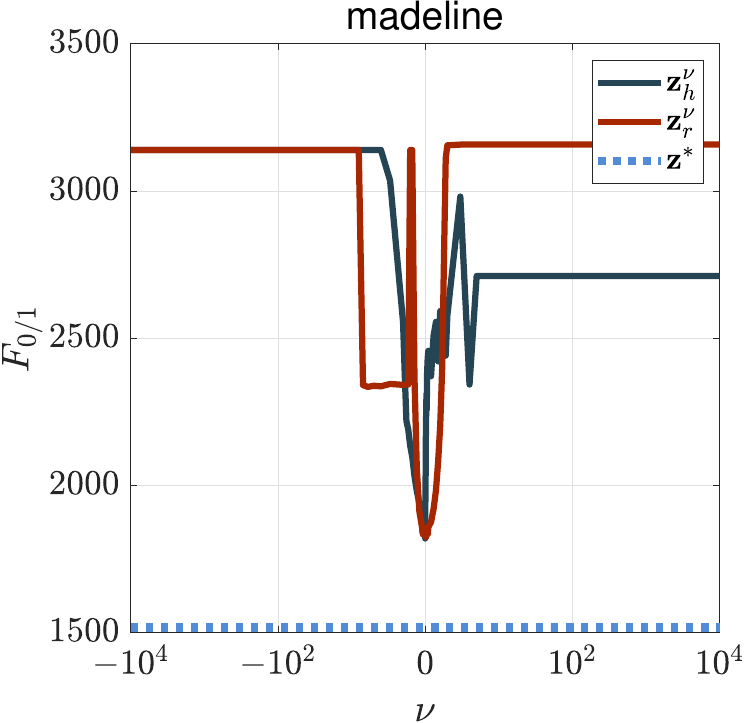}
			\end{minipage}%
			\begin{minipage}[t]{0.25\linewidth}
				\centering
				\includegraphics[width=1.6in]{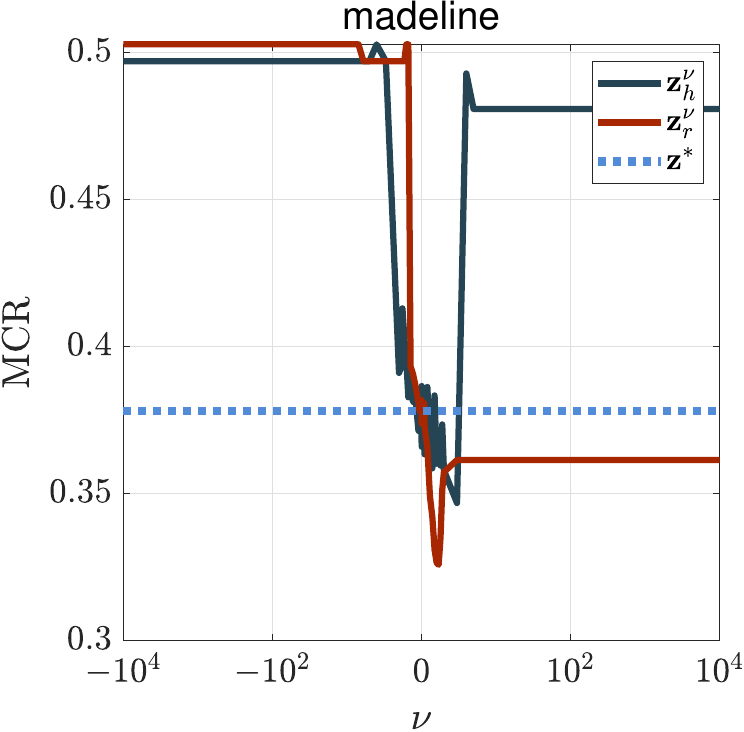}
			\end{minipage}
			\begin{minipage}[t]{0.25\linewidth}
				\centering
				\includegraphics[width=1.6in]{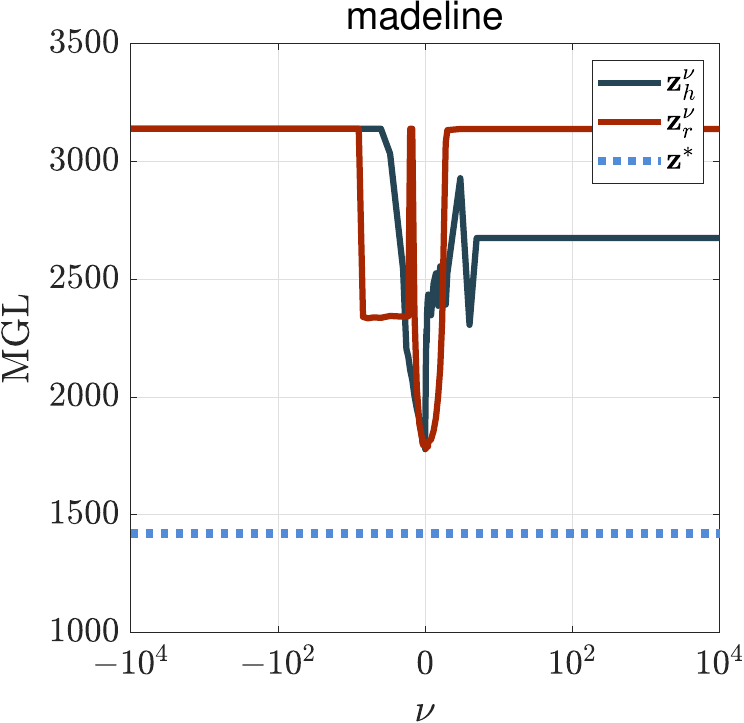}
		\end{minipage}}
		\caption{{Numerical results on dataset \texttt{madeline}.}} \label{fig-madeline}
		{}
	\end{figure*}

\begin{figure*}[htb]
		\subfloat{
			\begin{minipage}[t]{0.25\linewidth}
				\centering
				\includegraphics[width=1.68in]{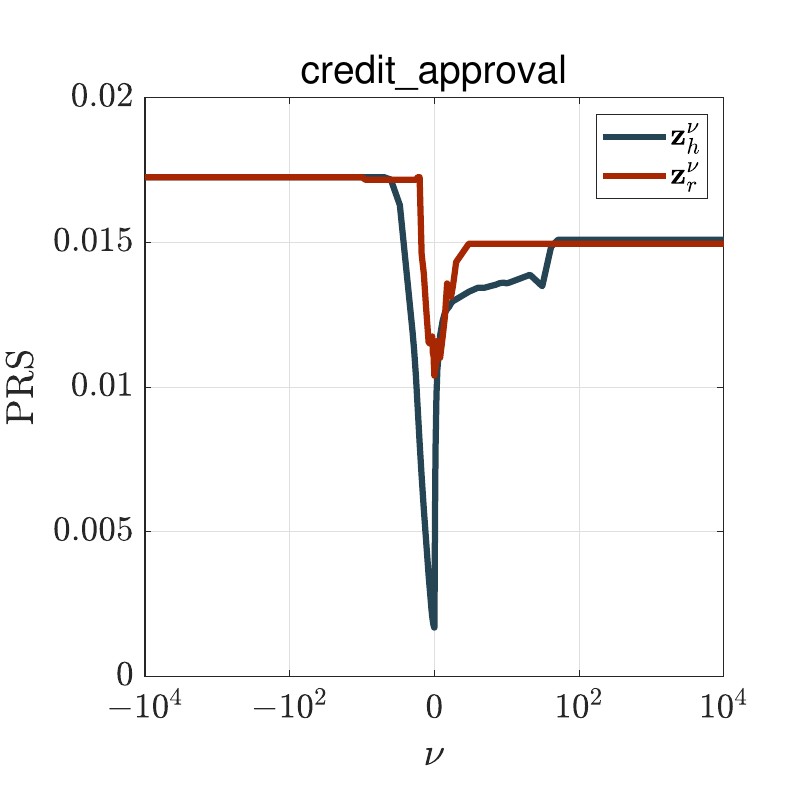}
			\end{minipage}%
			\begin{minipage}[t]{0.25\linewidth}
				\centering
				\includegraphics[width=1.68in]{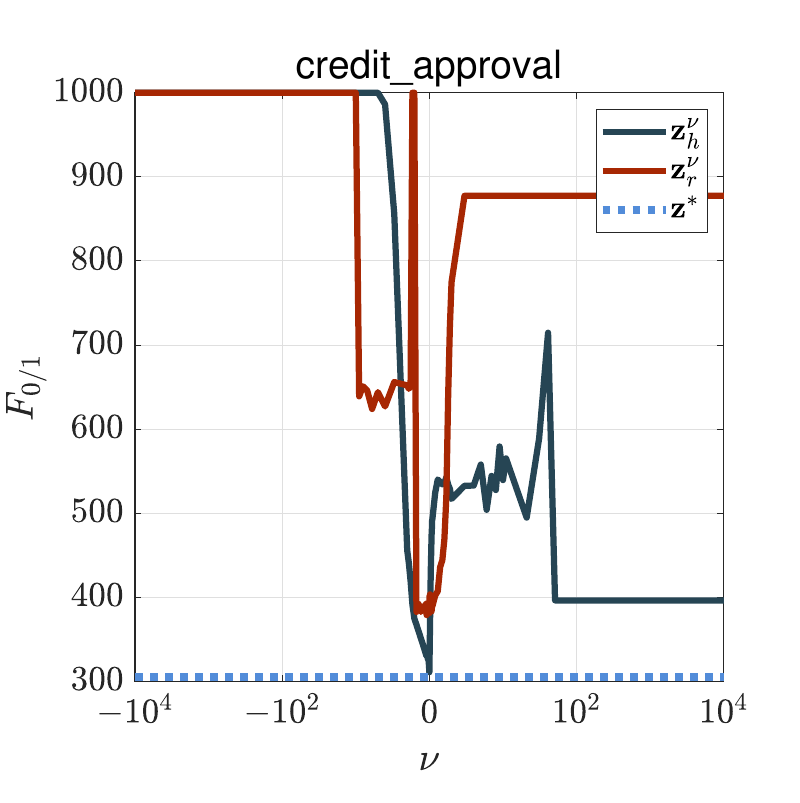}
			\end{minipage}%
			\begin{minipage}[t]{0.25\linewidth}
				\centering
				\includegraphics[width=1.66in]{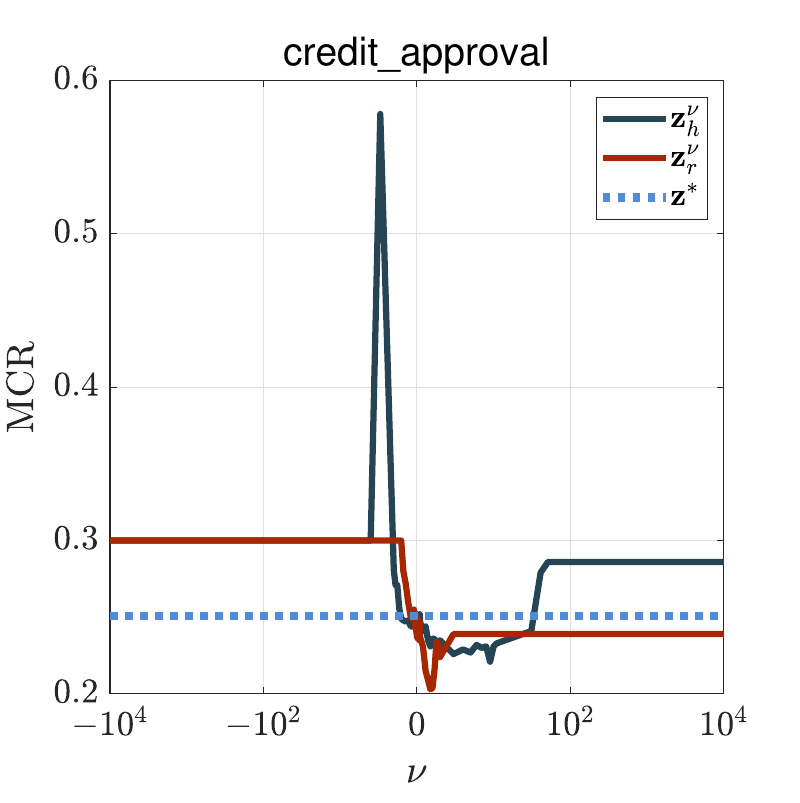}
			\end{minipage}
			\begin{minipage}[t]{0.25\linewidth}
				\centering
				\includegraphics[width=1.68in]{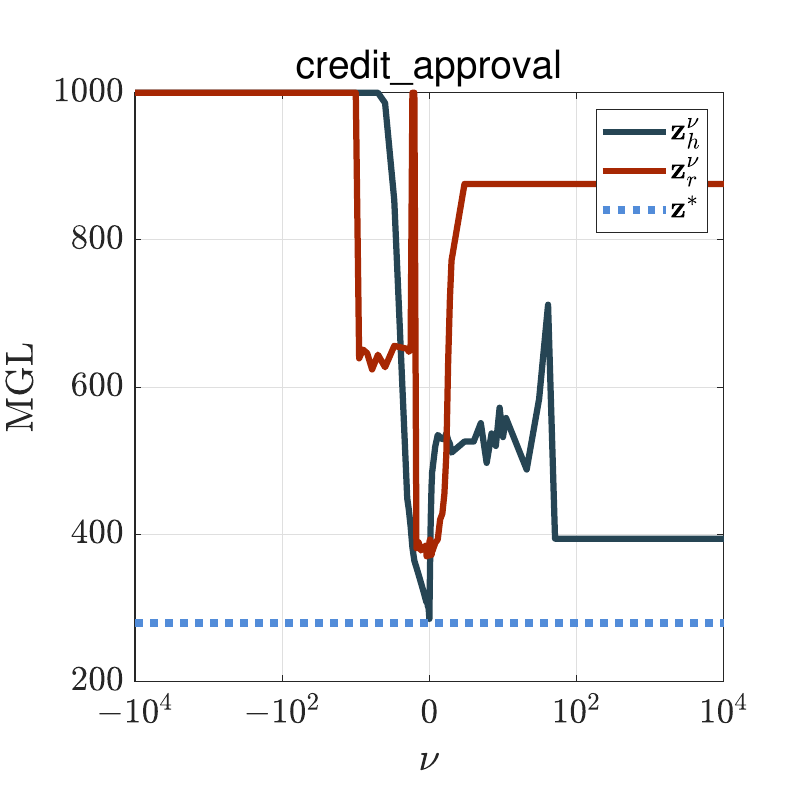}
		\end{minipage}}
		\caption{{Numerical results on dataset \texttt{credit\_approval}.}} \label{fig-Credit_Approval_Classification}
		{}
	\end{figure*}

\begin{figure*}[htb]
		\subfloat{
			\begin{minipage}[t]{0.25\linewidth}
				\centering
				\includegraphics[width=1.60in]{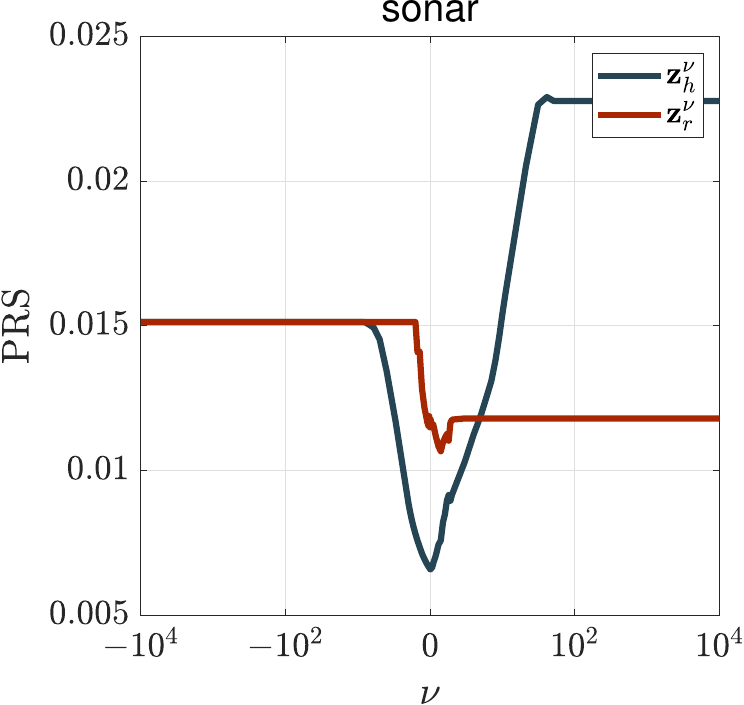}
			\end{minipage}%
			\begin{minipage}[t]{0.25\linewidth}
				\centering
				\includegraphics[width=1.54in]{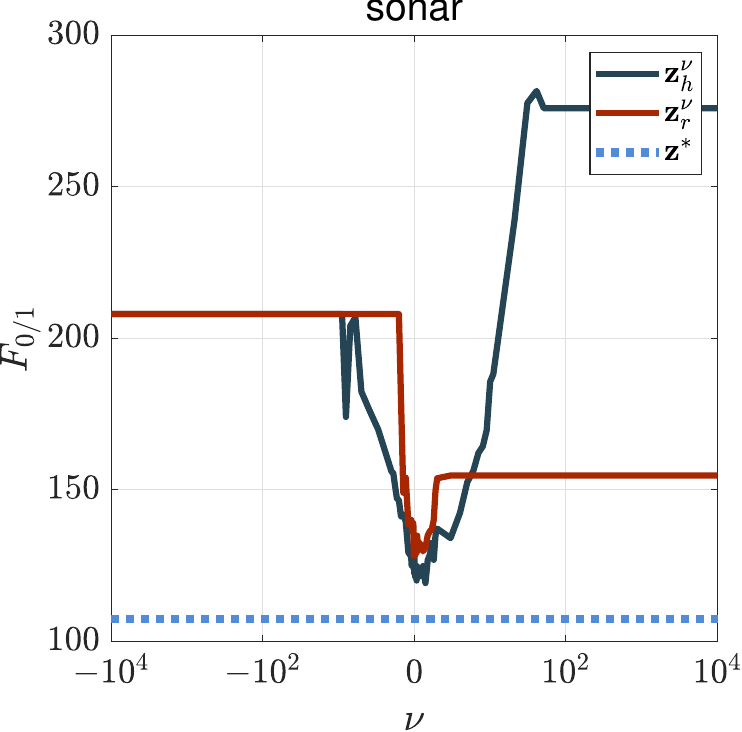}
			\end{minipage}%
			\begin{minipage}[t]{0.25\linewidth}
				\centering
				\includegraphics[width=1.54in]{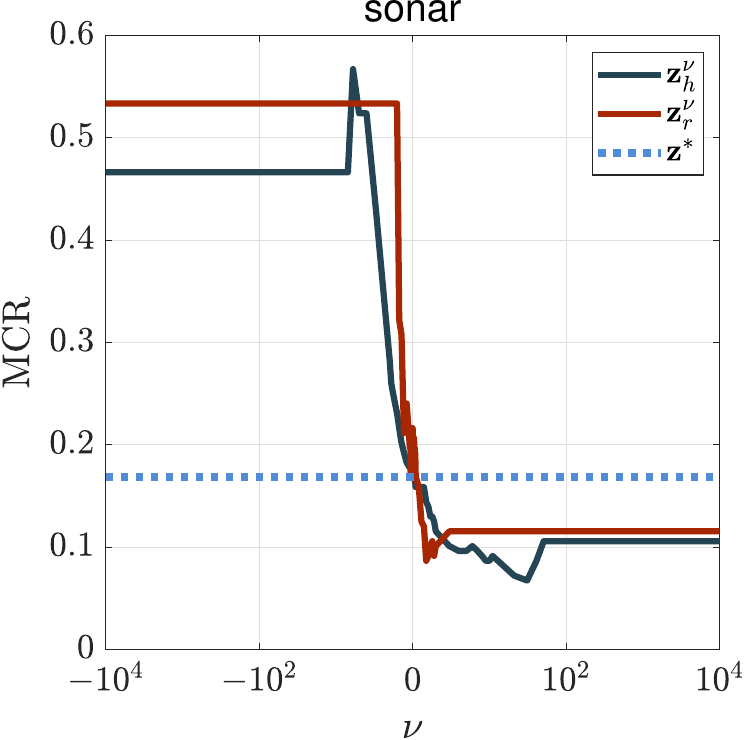}
			\end{minipage}
			\begin{minipage}[t]{0.25\linewidth}
				\centering
				\includegraphics[width=1.54in]{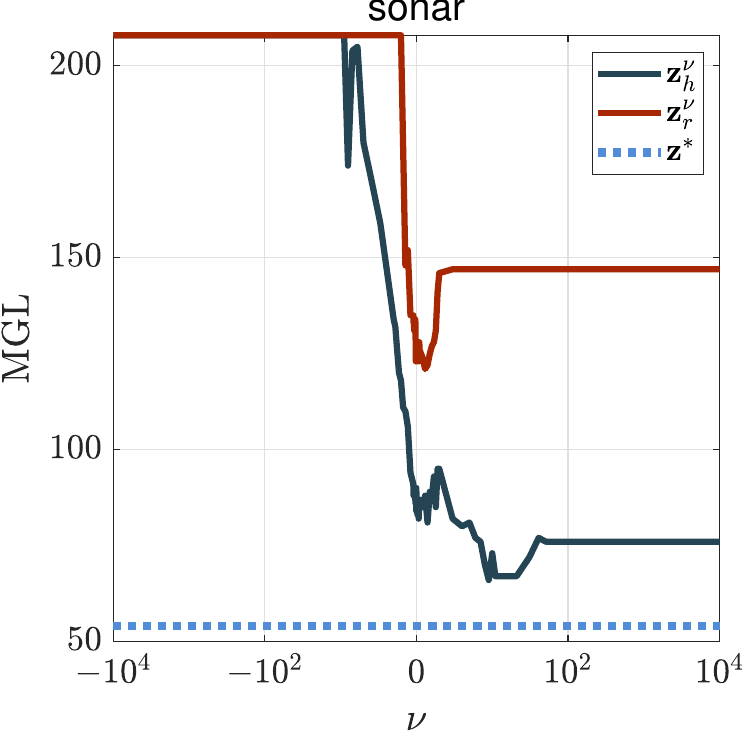}
		\end{minipage}}
		\caption{{Numerical results on dataset \texttt{sonar}.}} \label{fig-sonar}
		{}
	\end{figure*}

\begin{figure*}[htb]
		\subfloat{
			\begin{minipage}[t]{0.25\linewidth}
				\centering
				\includegraphics[width=1.6in]{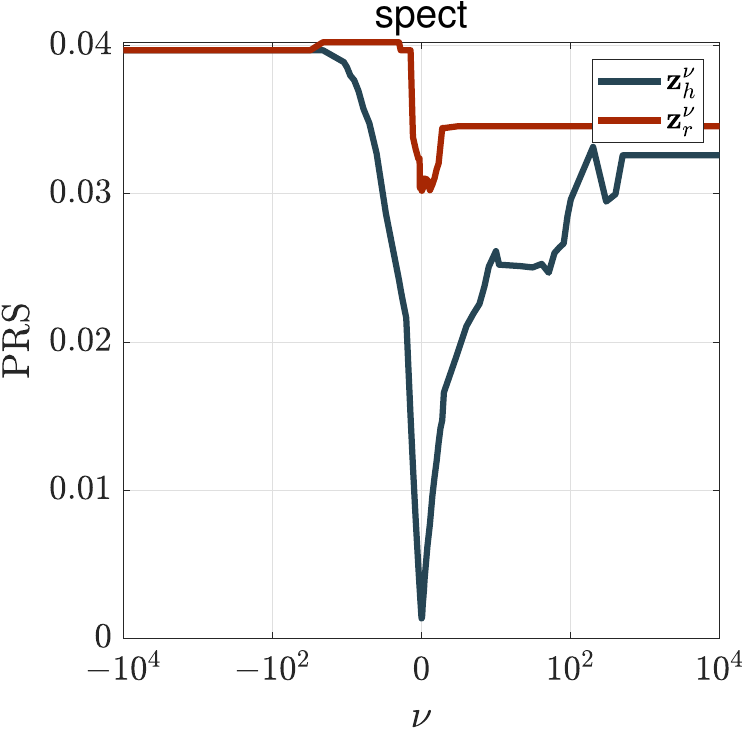}
			\end{minipage}%
			\begin{minipage}[t]{0.25\linewidth}
				\centering
				\includegraphics[width=1.57in]{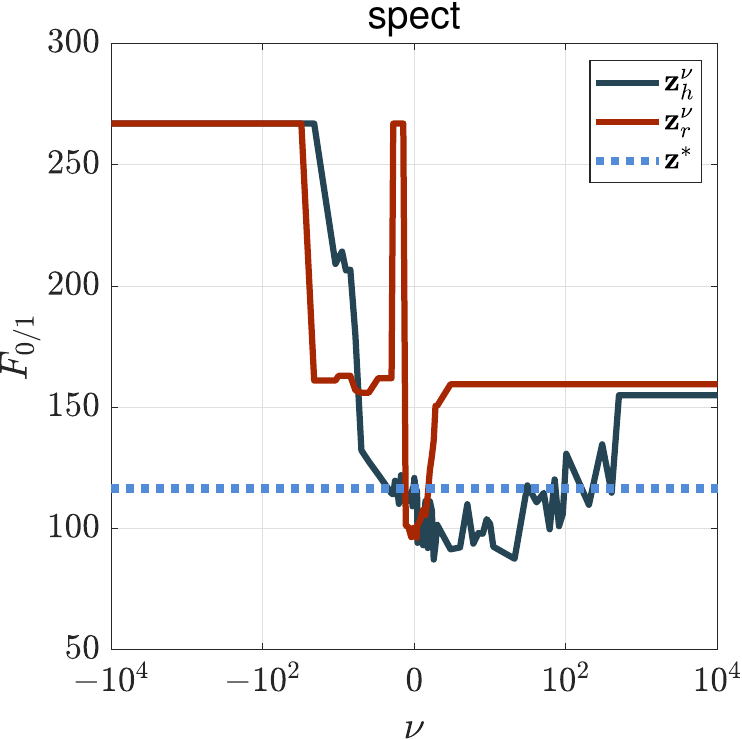}
			\end{minipage}%
			\begin{minipage}[t]{0.25\linewidth}
				\centering
				\includegraphics[width=1.57in]{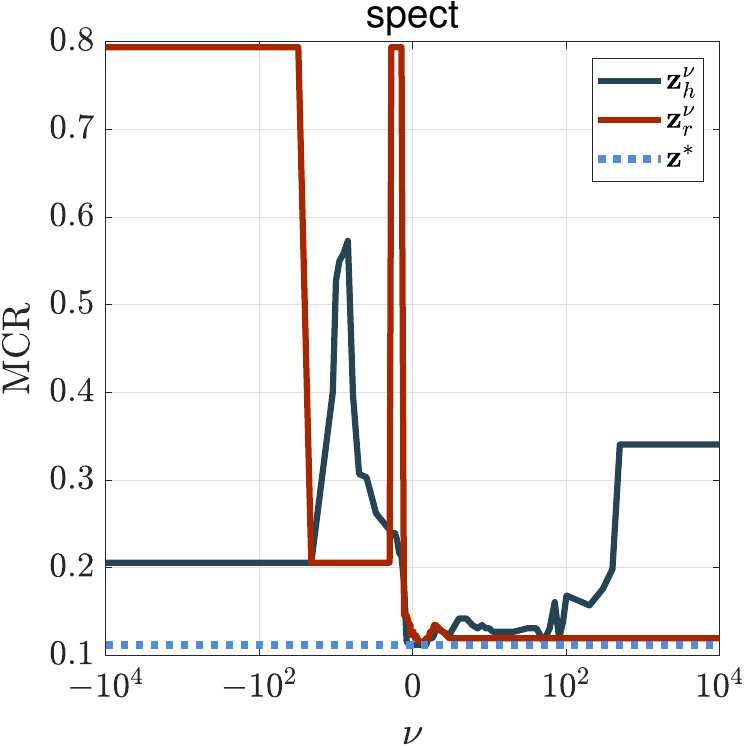}
			\end{minipage}
			\begin{minipage}[t]{0.25\linewidth}
				\centering
				\includegraphics[width=1.57in]{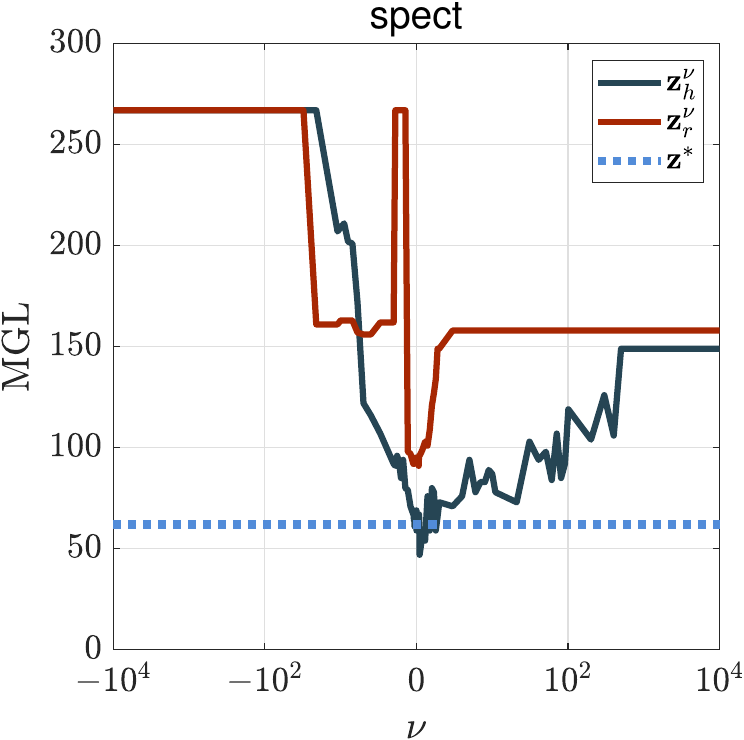}
		\end{minipage}}
		\caption{{Numerical results on dataset \texttt{spect}.}} \label{fig-spect}
		{}
	\end{figure*}

\begin{figure*}[htb]
		\subfloat{
			\begin{minipage}[t]{0.25\linewidth}
				\centering
				\includegraphics[width=1.61in]{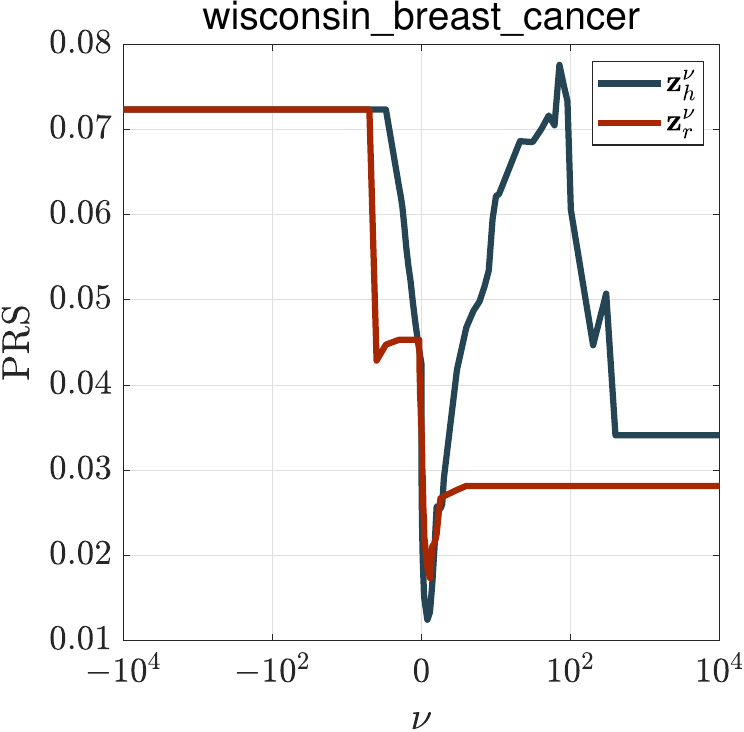}
			\end{minipage}%
			\begin{minipage}[t]{0.25\linewidth}
				\centering
				\includegraphics[width=1.6in]{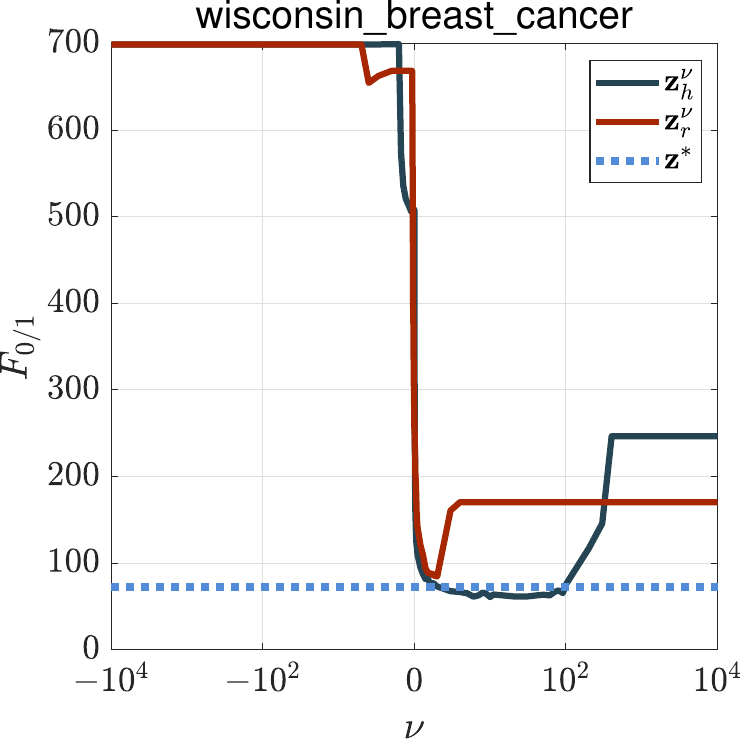}
			\end{minipage}%
			\begin{minipage}[t]{0.25\linewidth}
				\centering
				\includegraphics[width=1.6in]{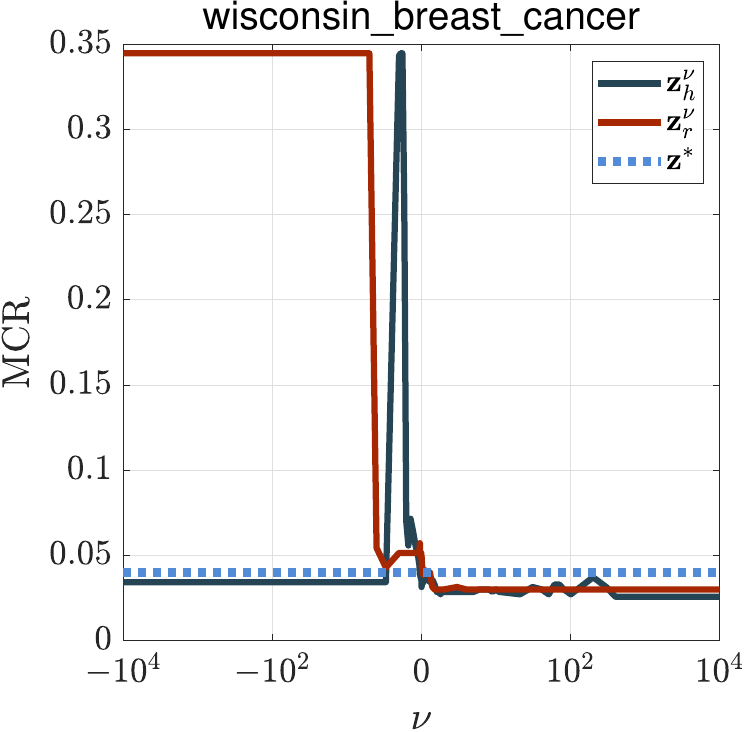}
			\end{minipage}
			\begin{minipage}[t]{0.25\linewidth}
				\centering
				\includegraphics[width=1.6in]{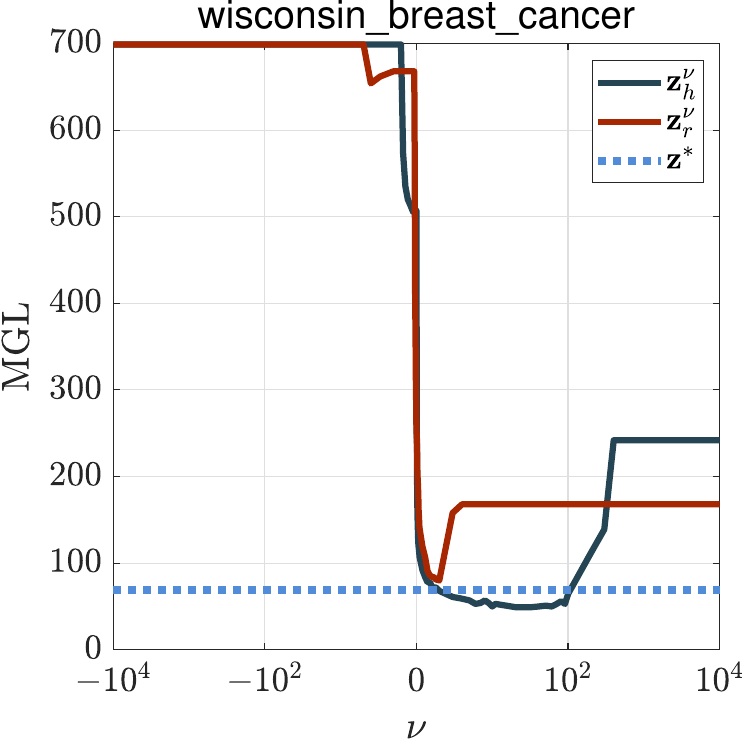}
		\end{minipage}}
		\caption{{Numerical results on dataset \texttt{wisconsin\_breast\_cancer}.}} \label{fig-wisconsin_breast_cancer}
		{}
	\end{figure*}

	We will record the following evaluation metrics with $\bfz$ taken from $\{ \bfz^*, \bfz^\nu_h, \bfz^\nu_r \}$:
	\begin{align*}
		& \mbox{Primal Residuals (PRS)}:= \frac{\| \bfz - \bfz^* \|}{(1+n)(1+\| \bfz^* \|)}, \\
		& \mbox{Primal Objective Value ($F_{0/1}$)} :=  \frac{1}{2} \| \bfw \|^2 + \lambda\sum_{i = 1}^m \ell_{0/1} \big( 1 - y_i (\langle \bfw, \bfx_i \rangle + b) \big), \\
		& \mbox{Misclassification Rate (MCR)} :=  1 - \frac{1}{m} \sum_{i = 1}^m \ell_{0/1} \big(y_i (\langle \bfw, \bfx_i \rangle + b) \big). \\
		& {\mbox{Margin Loss (MGL)} :=  \sum_{i = 1}^m \ell_{0/1} \big( 1 - y_i (\langle \bfw, \bfx_i \rangle + b) \big).}
	\end{align*}
	Here PRS reflects the relative residual between a point $\bfz$ to the local solution $\bfz^*$. This metric helps us verify the convergence of $\bfz^\nu_h$ to $\bfz^*$ in Theorem \ref{thm-p0-sequential-p1}. $F_{0/1}$ is the objective function value of \eqref{P01} and we set $\lambda = 1$ in experiments. MCR directly reflects the performance of classification on the given dataset. 
	{MGL is an upper bound to control the expectation of misclassification rate, and it helps to evaluate the generalization performance of a classifier (see \cite[Theorem 2, Corollary 6]{kakade2008complexity}).}  A smaller value of these four metrics indicates a better solution. Fig. \ref{fig-a1a}-\ref{fig-wisconsin_breast_cancer} demonstrates the performance of local solution $\bfz^*$ of \eqref{P01}, global solutions $\bfz^\nu_h$ of \eqref{P1}, and local solutions $\bfz^\nu_r$ of \eqref{Pr}.
	
			Finally, let us give some comments based on the observation of Figs. \ref{fig-a1a}-\ref{fig-wisconsin_breast_cancer}.
			
			$\bullet$ PRS: When $\nu$ is close to $0$, hyperparameters $\bfc^\nu$, $\rho^\nu$, and $\gamma^\nu$ get close to $\bfc$, $\rho$, and $\gamma$ respectively. 
			{As a result, PRS at $\bfz^\nu_h$ and $\bfz^\nu_r$ both drop significantly, and achieve almost the smallest values at $\nu = 0$. This means that $\bfz^0_h$ and $\bfz^0_r$ are near $\bfz^*$. Thms. \ref{thm-p0-sequential-p1} and \ref{theorem-P01-Pr} provide possible explanation for these observations.}
			
			$\bullet$ $F_{0/1}$: 
			{The point $\bfz^*$ is only a local solution of \eqref{P01}, but $F_{0/1}(\bfz^*)$ is smaller than $F_{0/1}(\bfz^\nu_h)$ and $F_{0/1}(\bfz^\nu_r)$ for most $\nu \in [-10^4,10^4]$.} Another interesting observation is that even though $\bfz^0_h$ (or $\bfz^0_r$) is very close to $\bfz^*$ (from PRS), there may still be a gap between $F_{0/1}(\bfz^*)$ and $F_{0/1}(\bfz^0_h)$ (or $F_{0/1}(\bfz^0_r)$); see, e.g. Figs. \ref{fig-a1a} and \ref{fig-madeline}. This is because the function $F_{0/1}$ is discontinuous.
			
			$\bullet$ MCR: On the datasets \texttt{a1a} and \texttt{heart\_scale}, MCR at $\bfz^*$ is smaller than that at $\bfz^\nu $ for most $\nu \in [-10^4,10^4]$. {However, MCR at $\bfz^*$ is larger than that of $\bfz^\nu_h$ or $\bfz^\nu_r$ for $\nu$ in specific ranges (e.g. datasets \texttt{german}, \texttt{madeline}). The main reason is that the 0/1-loss SVM \eqref{P01} includes the minimization of MGL and $\ell_2$ regularization term to improve the generalization performance of the classifier. Particularly, MGL counts the number of samples satisfying $y_i (\langle \bfw, \bfx_i \rangle + b) < 1$, whereas MCR measures the fraction of misclassified samples with $y_i (\langle \bfw, \bfx_i \rangle + b) \leq 0$ for $i \in [m]$. Therefore, the local minimizer $\bfz^*$ sometimes does not directly leads to a smaller MCR in practice.}
			
			{$\bullet$ MGL: We can observe that MGL at $\bfz^*$ is also smaller than that at $\bfz^\nu_h$ or $\bfz^\nu_r$ for most $\nu \in [-10^{-4}, 10^4]$. This observation verifies that 0/1-loss SVM \eqref{P01} optimizes MGL, and meanwhile suggests that $\bfz^*$ is possible to have better generalization performance.} 
			
			%
			%

			%
			%

			\section{Conclusions} \label{Section-Conclusion}
			
			The $\ell_0$-norm driven SSVM has been considered as an efficient model for dealing with large-scale classification, but there is lacking theoretical justification. This paper solves several fundamental issues of an SSVM formulation, which is the dual problem of 0/1-loss SVM.
			We note that the $\ell_0$-norm naturally arises from the dual problem and
			local duality is established  between SSVM and 0/1-loss SVM.
			Particularly, the linear representer theorem holds for their locally nice solutions. 
			Furthermore, the relationship between local solutions of 0/1-loss SVM,
			rSVM, and global solutions of convex SVMs is explored. This provides theoretical evidence for the satisfactory performance of 0/1-loss SVM and SSVM in comparison with convex SVMs. 
			Our results also justify the wide use of $\ell_0$-norm as a sparsity-driven tool.
			An important issue to be investigated is how to solve \eqref{D01} efficiently. This nonconvex problem has a different structure to other SVM formulations. Precisely tailored algorithms need to be devised for it in future work.
\bibliographystyle{IEEEtran}
\bibliography{01SVM_ref} 

\begin{thebibliography}{10}
\providecommand{\url}[1]{#1}
\csname url@samestyle\endcsname
\providecommand{\newblock}{\relax}
\providecommand{\bibinfo}[2]{#2}
\providecommand{\BIBentrySTDinterwordspacing}{\spaceskip=0pt\relax}
\providecommand{\BIBentryALTinterwordstretchfactor}{4}
\providecommand{\BIBentryALTinterwordspacing}{\spaceskip=\fontdimen2\font plus
\BIBentryALTinterwordstretchfactor\fontdimen3\font minus
  \fontdimen4\font\relax}
\providecommand{\BIBforeignlanguage}[2]{{%
\expandafter\ifx\csname l@#1\endcsname\relax
\typeout{** WARNING: IEEEtran.bst: No hyphenation pattern has been}%
\typeout{** loaded for the language `#1'. Using the pattern for}%
\typeout{** the default language instead.}%
\else
\language=\csname l@#1\endcsname
\fi
#2}}
\providecommand{\BIBdecl}{\relax}
\BIBdecl

\bibitem{tillmann2024cardinality}
A.~M. Tillmann, D.~Bienstock, A.~Lodi, and A.~Schwartz, ``Cardinality
  minimization, constraints, and regularization: a survey,'' \emph{SIAM Rev.},
  vol.~66, no.~3, pp. 403--477, 2024.

\bibitem{nikolova2013description}
M.~Nikolova, ``Description of the minimizers of least squares regularized with
  $\ell_0$-norm. {U}niqueness of the global minimizer,'' \emph{SIAM J. Imaging
  Sci.}, vol.~6, no.~2, pp. 904--937, 2013.

\bibitem{akkaya2020minimizers}
D.~Akkaya and M.~{\c{C}}. P{\i}nar, ``Minimizers of sparsity regularized huber
  loss function,'' \emph{J. Optim. Theory Appl.}, vol. 187, no.~1, pp.
  205--233, 2020.

\bibitem{akkaya2025minimizers}
------, ``Minimizers of sparsity regularized least absolute deviations,''
  \emph{J. Glob. Optim.}, pp. 1--27, 2025.

\bibitem{cortes1995support}
C.~Cortes and V.~Vapnik, ``Support-vector networks,'' \emph{Mach. Learn.},
  vol.~20, pp. 273--297, 1995.

\bibitem{vapnik1998statistical}
V.~Vapnik, ``Statistical learning theory,'' \emph{New York}, 1998.

\bibitem{cristianini2000introduction}
N.~Cristianini and J.~Shawe-Taylor, \emph{An introduction to support vector
  machines and other kernel-based learning methods}.\hskip 1em plus 0.5em minus
  0.4em\relax Cambridge university press, 2000.

\bibitem{steinwart2008support}
I.~Steinwart and A.~Christmann, ``Support vector machines,'' 2008.

\bibitem{campbell2011learning}
C.~Campbell and Y.~Ying, \emph{Learning with support vector machines}.\hskip
  1em plus 0.5em minus 0.4em\relax Morgan \& Claypool Publishers, 2011.

\bibitem{weston2003use}
J.~Weston, A.~Elisseeff, B.~Sch{\"o}lkopf, and M.~Tipping, ``Use of the
  zero-norm with linear models and kernel methods,'' \emph{J. Mach. Learn.
  Res.}, vol.~3, no. Mar, pp. 1439--1461, 2003.

\bibitem{guan2009mixed}
W.~Guan, A.~Gray, and S.~Leyffer, ``Mixed-integer support vector machine,'' in
  \emph{NIPS workshop on optimization for machine learning}, 2009.

\bibitem{fung2002minimal}
G.~M. Fung, O.~L. Mangasarian, and A.~J. Smola, ``Minimal kernel classifiers,''
  \emph{J. Mach. Learn. Res.}, vol.~3, no. Nov, pp. 303--321, 2002.

\bibitem{chan2007direct}
A.~B. Chan, N.~Vasconcelos, and G.~R. Lanckriet, ``Direct convex relaxations of
  sparse {SVM},'' in \emph{Proc. 24th Int. Conf. Mach. Learn.}, 2007, pp.
  145--153.

\bibitem{subrahmanya2009sparse}
N.~Subrahmanya and Y.~C. Shin, ``Sparse multiple kernel learning for signal
  processing applications,'' \emph{IEEE Trans. Pattern Anal. Mach. Intell.},
  vol.~32, no.~5, pp. 788--798, 2009.

\bibitem{shao2018sparse}
Y.~Shao, C.~Li, M.~Liu, Z.~Wang, and N.~Deng, ``Sparse $l_q$-norm least squares
  support vector machine with feature selection,'' \emph{Pattern Recogn.},
  vol.~78, pp. 167--181, 2018.

\bibitem{bomze2025feature}
I.~Bomze, F.~D’Onofrio, L.~Palagi, and B.~Peng, ``Feature selection in linear
  support vector machines via a hard cardinality constraint: a scalable conic
  decomposition approach,'' \emph{Eur. J. Oper. Res.}, 2025.

\bibitem{zhang2025sparse}
P.~Zhang, N.~Xiu, and H.-D. Qi, ``Sparse {SVM} with hard-margin loss: a
  {N}ewton-augmented {L}agrangian method in reduced dimensions,'' \emph{J.
  Mach. Learn. Res.}, vol.~26, no. 118, pp. 1--55, 2025.

\bibitem{dinuzzo2012representer}
F.~Dinuzzo and B.~Sch{\"o}lkopf, ``The representer theorem for {H}ilbert
  spaces: a necessary and sufficient condition,'' in \emph{Proc. Adv. Neural
  Inf. Process. Syst}, vol.~25, 2012.

\bibitem{zhou2021sparse}
S.~Zhou, ``Sparse {SVM} for sufficient data reduction,'' \emph{IEEE Trans.
  Pattern Anal. Mach. Intell.}, vol.~44, no.~9, pp. 5560--5571, 2021.

\bibitem{wang2021support}
H.~Wang, Y.~Shao, S.~Zhou, C.~Zhang, and N.~Xiu, ``Support vector machine
  classifier via $\ell_{0/1}$ soft-margin loss,'' \emph{IEEE Trans. Pattern
  Anal. Mach. Intell.}, vol.~44, no.~10, pp. 7253--7265, 2021.

\bibitem{girosi1998equivalence}
F.~Girosi, ``An equivalence between sparse approximation and support vector
  machines,'' \emph{Neural Comput.}, vol.~10, no.~6, pp. 1455--1480, 1998.

\bibitem{steinwart2003sparseness-nips}
I.~Steinwart, ``Sparseness of support vector machines---some asymptotically
  sharp bounds,'' \emph{Proc. Adv. Neural Inf. Process. Syst.}, vol.~16, 2003.

\bibitem{huang2009arbitrary}
K.~Huang, D.~Zheng, I.~King, and M.~R. Lyu, ``Arbitrary norm support vector
  machines,'' \emph{Neural Comput.}, vol.~21, no.~2, pp. 560--582, 2009.

\bibitem{bi2003dimensionality}
J.~Bi, K.~Bennett, M.~Embrechts, C.~Breneman, and M.~Song, ``Dimensionality
  reduction via sparse support vector machines,'' \emph{J. Mach. Learn. Res.},
  vol.~3, no. Mar, pp. 1229--1243, 2003.

\bibitem{shao2019joint}
Y.~Shao, C.~Li, L.~Huang, Z.~Wang, N.~Deng, and Y.~Xu, ``Joint sample and
  feature selection via sparse primal and dual {LSSVM},'' \emph{Knowl.-Based
  Syst.}, vol. 185, p. 104915, 2019.

\bibitem{huang2010sparse}
K.~Huang, D.~Zheng, J.~Sun, Y.~Hotta, K.~Fujimoto, and S.~Naoi, ``Sparse
  learning for support vector classification,'' \emph{Pattern Recogn. Lett.},
  vol.~31, no.~13, pp. 1944--1951, 2010.

\bibitem{lopez2011sparse}
J.~Lopez, K.~De~Brabanter, J.~Dorronsoro, and J.~Suykens, ``Sparse {LS-SVM}s
  with $\ell_0$-norm minimization,'' in \emph{Proc. Eur. Symp. Artif. Neural
  Netw., Comput. Intell. Mach. Learn.}, 2011, pp. 189--194.

\bibitem{landeros2023algorithms}
A.~Landeros and K.~Lange, ``Algorithms for sparse support vector machines,''
  \emph{J. Comput. Graph. Stat.}, vol.~32, no.~3, pp. 1097--1108, 2023.

\bibitem{wu2007robust}
Y.~Wu and Y.~Liu, ``Robust truncated hinge loss support vector machines,''
  \emph{J. Am. Stat. Assoc.}, vol. 102, no. 479, pp. 974--983, 2007.

\bibitem{brooks2011support}
J.~P. Brooks, ``Support vector machines with the ramp loss and the hard margin
  loss,'' \emph{Oper. Res.}, vol.~59, no.~2, pp. 467--479, 2011.

\bibitem{cotter2013learning}
A.~Cotter, S.~Shalev-Shwartz, and N.~Srebro, ``Learning optimally sparse
  support vector machines,'' in \emph{Proc. Int. Conf. Mach. Learn.}, 2013, pp.
  266--274.

\bibitem{collobert2006trading}
R.~Collobert, F.~Sinz, J.~Weston, and L.~Bottou, ``Trading convexity for
  scalability,'' in \emph{Proc. 23rd Int. Conf. Mach. Learn.}, 2006, pp.
  201--208.

\bibitem{ong2013learning}
C.~S. Ong and L.~T.~H. An, ``Learning sparse classifiers with difference of
  convex functions algorithms,'' \emph{Optim. Methods Softw.}, vol.~28, no.~4,
  pp. 830--854, 2013.

\bibitem{li2007optimizing}
L.~Li and H.-T. Lin, ``Optimizing 0/1 loss for perceptrons by random coordinate
  descent,'' in \emph{Proc. Int. Joint Conf. Neural Netw.}\hskip 1em plus 0.5em
  minus 0.4em\relax IEEE, 2007, pp. 749--754.

\bibitem{zhang2025composite}
P.~Zhang, N.~Xiu, and H.~Qi, ``Composite optimization with indicator functions:
  stationary duality and a semismooth {N}ewton method,'' \emph{Math. Program.},
  pp. 1--46, 2025.

\bibitem{shawe2004kernel}
J.~Shawe-Taylor and N.~Cristianini, \emph{Kernel methods for pattern
  analysis}.\hskip 1em plus 0.5em minus 0.4em\relax Cambridge University Press,
  2004.

\bibitem{burges1999uniqueness}
C.~J. Burges and D.~Crisp, ``Uniqueness of the {SVM} solution,'' in \emph{Proc.
  Adv. Neural Inf. Process. Syst.}, vol.~12, 1999.

\bibitem{lapin2014learning}
M.~Lapin, M.~Hein, and B.~Schiele, ``Learning using privileged information:
  {SVM}+ and weighted {SVM},'' \emph{Neural Netw.}, vol.~53, pp. 95--108, 2014.

\bibitem{RockWets98}
R.~T. Rockafellar and R.~J.-B. Wets, \emph{Variational Analysis}.\hskip 1em
  plus 0.5em minus 0.4em\relax Springer, Berlin, 1998.

\bibitem{rockafellar1970convex}
R.~T. Rockafellar, \emph{Convex Analysis}.\hskip 1em plus 0.5em minus
  0.4em\relax Princeton University Press, 1970.

\bibitem{karasuyama2012multi}
M.~Karasuyama, N.~Harada, M.~Sugiyama, and I.~Takeuchi, ``Multi-parametric
  solution-path algorithm for instance-weighted support vector machines,''
  \emph{Mach. Learn.}, vol.~88, no.~3, pp. 297--330, 2012.

\bibitem{iranmehr2019cost}
A.~Iranmehr, H.~Masnadi-Shirazi, and N.~Vasconcelos, ``Cost-sensitive support
  vector machines,'' \emph{Neurocomputing}, vol. 343, pp. 50--64, 2019.

\bibitem{nocedal2006numerical}
J.~Nocedal and S.~Wright, \emph{Numerical optimization}.\hskip 1em plus 0.5em
  minus 0.4em\relax Springer, New York, 2006.

\bibitem{kakade2008complexity}
S.~M. Kakade, K.~Sridharan, and A.~Tewari, ``On the complexity of linear
  prediction: Risk bounds, margin bounds, and regularization,'' \emph{Proc.
  Adv. Neural Inf. Process. Syst.}, vol.~21, 2008.

\bibitem{nguyen2013algorithms}
T.~Nguyen and S.~Sanner, ``Algorithms for direct 0--1 loss optimization in
  binary classification,'' in \emph{Proc. Int. Conf. Mach. Learn.}\hskip 1em
  plus 0.5em minus 0.4em\relax PMLR, 2013, pp. 1085--1093.

\bibitem{zhou2021global}
S.~Zhou, N.~Xiu, and H.-D. Qi, ``Global and quadratic convergence of {N}ewton
  hard-thresholding pursuit,'' \emph{J. Mach. Learn. Res.}, vol.~22, no.~12,
  pp. 1--45, 2021.

\bibitem{zhou2021quadratic}
S.~Zhou, L.~Pan, N.~Xiu, and H.-D. Qi, ``Quadratic convergence of smoothing
  {N}ewton's method for 0/1 loss optimization,'' \emph{SIAM J. Optim.},
  vol.~31, no.~4, pp. 3184--3211, 2021.

\bibitem{chang2011libsvm}
C.-C. Chang and C.-J. Lin, ``{LIBSVM}: a library for support vector machines,''
  \emph{ACM Trans. Intell. Syst. Technol.}, vol.~2, no.~3, pp. 1--27, 2011.

\end{thebibliography}

\end{document}